\documentclass[journal]{IEEEtran}
\IEEEoverridecommandlockouts
\usepackage{graphicx}
\def\BibTeX{{\rm B\kern-.05em{\sc i\kern-.025em b}\kern-.08em
    T\kern-.1667em\lower.7ex\hbox{E}\kern-.125emX}}
\usepackage[font=small,labelfont=bf,tableposition=top]{caption}

\usepackage{blindtext}
\usepackage{caption}

\usepackage{amsmath,amssymb,amsfonts}
\usepackage[inline]{enumitem}
\usepackage{graphicx}
\usepackage{textcomp}

\usepackage{booktabs}
\usepackage{multirow}
\usepackage{hyperref}
\hypersetup{
colorlinks=true,linkcolor=red,citecolor=blue,urlcolor=black}

\usepackage[utf8]{inputenc}
\usepackage[T1]{fontenc}
\usepackage{float}
\usepackage{adjustbox}
\usepackage{array}
\usepackage{etoolbox}
\usepackage{dsfont}

\usepackage{cite}
\usepackage{amsmath,am            ssymb,amsfonts}
\usepackage{graphicx}
\usepackage{textcomp}
\usepackage{xcolor}
\usepackage{booktabs}
\usepackage{multirow}
\usepackage{float}
\usepackage[hypcap=false]{caption}
\usepackage{graphicx}
\usepackage[section]{placeins}
\usepackage{etoolbox,lipsum}
\usepackage{afterpage}
\usepackage{booktabs}
\usepackage{subcaption}
\usepackage{capt-of,etoolbox}
\usepackage[section]{placeins}
\usepackage{etoolbox,lipsum}
\usepackage{multirow, makecell}
\usepackage{multirow}
\usepackage{xcolor,colortbl}
\usepackage{soul}
\usepackage{color}
\usepackage{cancel}
\newcommand{\mypar}[1]{\vspace{0.3cm}\noindent\textbf{#1}}
\usepackage{expl3}
\ExplSyntaxOn
\newcommand\latinabbrev[1]{
  \peek_meaning:NTF . {
    #1\@}%
  { \peek_catcode:NTF a {
      #1.\@ }%
    {#1.\@}}}
\ExplSyntaxOff
\def\eg{\latinabbrev{e.g}}
\def\etal{\latinabbrev{et al}}

\def\ie{\latinabbrev{i.e}}

\usepackage{tikz}
\newcommand*\circled[1]{\tikz[baseline=(char.base)]{
		\node[shape=circle,draw,inner sep=0.2pt] (char) {#1};}}

\usepackage{balance}
\usepackage{stfloats}
\usepackage{bm}
\usepackage{esvect}
\definecolor{lightsteelblue}{RGB}{176,196,222}
\definecolor{lightsteelred}{RGB}{230,176,160}
\definecolor{lightsteellila}{RGB}{175,181,224}
\definecolor{lightsteelgreen}{RGB}{182,214,207}
\definecolor{lightsteelyellow}{RGB}{240,240,160}
\definecolor{lightsteelwhite}{RGB}{255,255,255}
\definecolor{lightsteelgray}{RGB}{205,201,201}
\definecolor{lightsteellightgray}{RGB}{210,210,210}

\definecolor{pp_blue}{RGB}{68,114,196}
\definecolor{pp_orange}{RGB}{237,125,49}
\definecolor{pp_lila}{RGB}{112,48,160}
\definecolor{pp_ygreen}{RGB}{112,173,71}

\definecolor{gg_blue}{RGB}{52,138,189}
\definecolor{gg_red}{RGB}{226,74,51}
\definecolor{gg_lila}{RGB}{152,142,213}
\definecolor{gg_green}{RGB}{112,173,71}

\definecolor{table_standard}{RGB}{230,153,0}
\definecolor{table_uncertainty}{RGB}{112,48,160}

\usepackage{tikz}

\usepackage{subcaption}

\usepackage[capitalise]{cleveref} 

\usepackage{listings}
\newcommand{\figref}[1]{\mbox{Figure~\ref{#1}}}

\newcommand{\revised}[1]{\textcolor{black}{#1}}

\usepackage{algorithm}

\begin{document}

\title{\textit{Is my \revised{Driver Observation Model} Overconfident?} \\\revised{Input-guided Calibration Networks for} Reliable and Interpretable Confidence Estimates}

\author{Alina Roitberg    \quad \quad Kunyu Peng  \quad \quad  David Schneider \\[5pt] \hspace{1cm} Kailun Yang   \quad\quad\quad Marios Koulakis  \quad\quad\quad  Manuel Martinez  \quad\quad\quad   Rainer Stiefelhagen
\vspace{0.5cm}
\thanks{This work was supported by the Competence Center Karlsruhe for AI Systems Engineering (CC-KING, \url{www.ai-engineering.eu}) sponsored by the
Ministry of Economic Affairs, Labour and Housing Baden-W{\"u}rttemberg.}
\thanks{Authors are with Institute for Anthropomatics and Robotics, Karlsruhe Institute of Technology, Germany (e-mail: firstname.lastname@kit.edu).}
\thanks{Corresponding author: Alina Roitberg.}
}

\makeatletter
\let\@oldmaketitle\@maketitle
\makeatother

\maketitle


\begin{abstract}

Driver observation models are rarely deployed under perfect conditions.
In practice, illumination, camera placement  and  type   differ from the ones present during training and unforeseen behaviours may occur at any time.
While observing the human behind the steering wheel leads to more intuitive human-vehicle-interaction and safer driving, it requires recognition algorithms which do not only predict the correct driver state, but also determine their prediction quality through \emph{realistic and interpretable confidence measures}.
Reliable uncertainty estimates are crucial for building trust and are a serious obstacle for deploying activity recognition networks in real driving systems.

In this work, we for the first time examine how well the confidence values of modern driver observation models indeed match the probability of the correct outcome and show that raw neural network-based approaches tend to significantly overestimate their prediction quality.
To correct this misalignment between the confidence values and the actual uncertainty, we consider two strategies. First, we enhance two activity recognition models often used for driver observation with  temperature scaling -- an off-the-shelf method for confidence calibration in image classification.
Then, we introduce Calibrated Action Recognition with Input Guidance (CARING) -- a novel approach leveraging an additional neural network to learn scaling the confidences \emph{depending on the video representation}.
Extensive experiments on the Drive\&Act dataset  demonstrate that both strategies drastically improve the quality of model confidences,  while our CARING model outperforms both, the original architectures and their temperature scaling enhancement, leading to best uncertainty estimates.

\end{abstract}


\IEEEpeerreviewmaketitle


\section{Introduction}

\begin{figure}
   \centering
   \includegraphics[width=\linewidth]{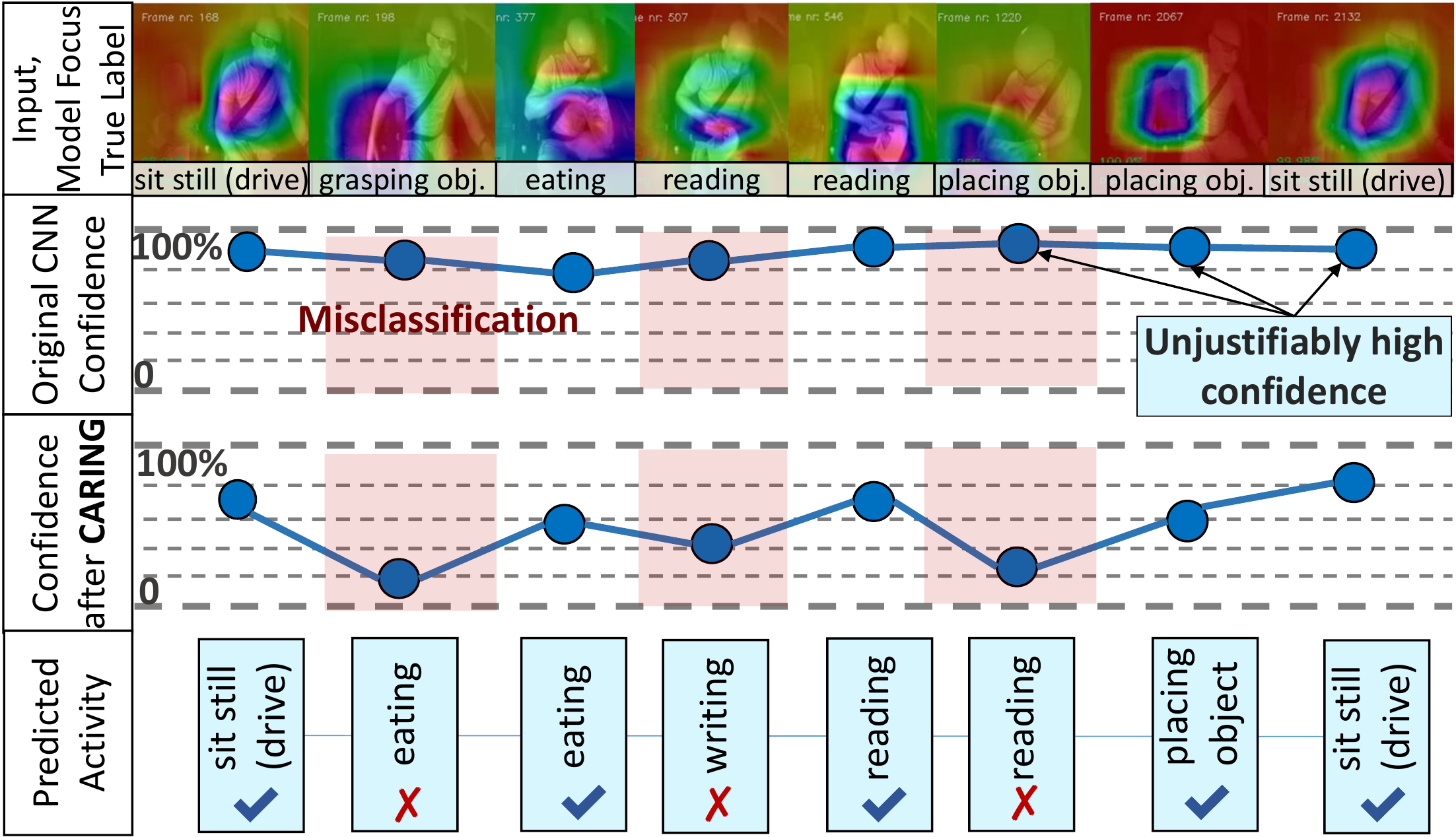}
   \caption{
   \revised{Deep neural networks have seen a surge in popularity in driver observation, but worrying evidence indicates that such models produce unjustifiably high-confidence predictions.
   Overconfident models imply that if a network predicts certain secondary behaviour with $>99\%$ confidence (which is indeed very common), the prediction is oftentimes correct in much less than in 99\% of cases.
   }
   \revised{This figure depicts an example of observed predictions and confidence values of a driver activity recognition CNN during a single driving session.} 
If we directly use the probability estimates provided by the original CNNs, we end up with \textit{unjustifiably high confidences}, which can be very damaging in practice. 
Temperature-based methods~\cite{guo2017calibration} apply a fixed scaling to the CNN probability estimates in order to obtain more realistic values.
Our approach, CARING, goes one step further, and dynamically infers a suitable scaling factor depending on the current observation, leading to more reliable confidences.
   }
   \label{fig:intro_gradcam}
\end{figure}

\IEEEPARstart{W}{ith} the rapidly growing accuracy of  driver observation models\cite{martin2019drive}, the existing gap between the published methods and their applications in practice makes us wonder about important performance aspects  potentially being  overlooked.
When examining the previous research, we make two observations\revised{. First, } the existing driver activity recognition algorithms are highly driven by the top-1 accuracy~\cite{hand_gesture,martin2018body,jain2015car,wharton2021coarse,weyers2019action,li2019novel,martin2019drive,gebert2019end}, skipping other relevant metrics, such as the \emph{reliability of their confidence values}\revised{. Second,} recent fundamental deep learning research raises alarm about neural networks being notably bad at detecting ambiguities~\cite{guo2017calibration,sunderhauf2018limits,hendrycks17baseline,nguyen2015deep}.
Despite the impressive gain in accuracy on modern driver activity recognition datasets~\cite{martin2019drive, wharton2021coarse, ortega2020dmd}, examining how well the confidence values of driver behaviour understanding models indeed reflect the probability of a correct prediction (see Figure \ref{fig:intro_gradcam}) has been overlooked in the past and is the main motivation of our work.

Humans have a natural intuition about probabilities~\cite{fontanari2014probabilistic}: 
If a neural network reports that the driver is ``talking on the phone'' with $99$\% confidence, we tend to  take this prediction for granted. 
In neural networks, the output of the last fully-connected is usually normalized through the \textit{Softmax}\footnote{\textit{Softmax} function is often applied on \textit{logits} (the output vector of the last layer of a classification network, where each value represents a category score) and normalizes them to sum up to one by computing the exponents of each output and then normalizing each of them by the sum of those exponents.}  function to sum up to one, therefore \emph{resembling} prediction probabilities.
However, blindly interpreting these values as probabilities of the outcome being correct  would be naive, as the Softmax scores are not calibrated and merely optimized for a high top-1 accuracy on a static set driver of behaviours. 
The output  Softmax vector  \textit{appears} to hold probabilities of the individual classes, but they do not necessarily match the true confidence of the model~\cite{gal2016dropout, guo2017calibration}. 
Furthermore, worrying evidence from deep learning research highlights the problem of \emph{model miscalibration}, suggesting that raw neural network confidences tend to be biased towards very high values for both, correct and incorrect predictions~\cite{guo2017calibration,gal2016dropout}. 
In real-life driving applications, on the other hand,  models need to be more reserved: wrong high-confidence decision can lead to tragic outcomes\revised{. At the same time,} indicating high classification uncertainty and reacting accordingly is the better choice. 
In this work, we equip driver observation frameworks with building blocks for reliably quantifying model confidence.

We aim to highlight the need for interpretable and reliable uncertainty measures as a means for identifying misclasifications inside the vehicle cabin\revised{. Although this area is seeing growing attention in the field of general image recognition~\cite{guo2017calibration,hendrycks17baseline}, it is underresearched in video classification and especially driver observation}.
To achieve this, we integrate the \emph{reliability} of model confidence in the large-scale driver activity recognition benchmark \emph{Drive\&Act}~\cite{martin2019drive} and develop models which learn to convert oftentimes overconfident uncertainty values of the original networks into reliable probability estimates.

 \looseness=-1

\mypar{Why driver activity recognition?} 
Rising levels of automation increase human freedom, leading to drivers being engaged in distractive behaviors more often while the type of activities become increasingly diverse.
For example, \textsl{working on laptop} or \textsl{reading magazine} behind the steering were almost unthinkable until now, but these behaviours become more common as the driver is gradually relieved from actively steering the car.
Although distractions become safer as the vehicle becomes more intelligent, this change does not happen from one day to another and is a rather long-lasting transformation~\cite{SAE-J3016}. 
Over-reliance on  artificial intelligence might lead to catastrophic consequences, and, for a long time, the driver will need to intervene in case of uncertainty~\cite{SAE-J3016,radlmayr2014traffic,martin_at}.
However, there are important long-term application scenarios of driver \revised{activity recognition} even in fully-autonomous cars.
For example, understanding the situation inside the vehicle cabin may increase comfort,  \eg, by adjusting the driving style if the person is drinking a hot drink or being an intuitive communication interface via gestures.
Applications of driver activity recognition models depend on the degree of vehicle automation~\cite{sae2014automated}.

We summarize four main use-cases for applications of driver activity recognition~\cite{flad2020personalisation,ohn2016looking,martin2019drive}: 
\begin{figure*}

  \begin{center}

\begin{subfigure}[b]{0.54\linewidth}
    \includegraphics[width=\linewidth]{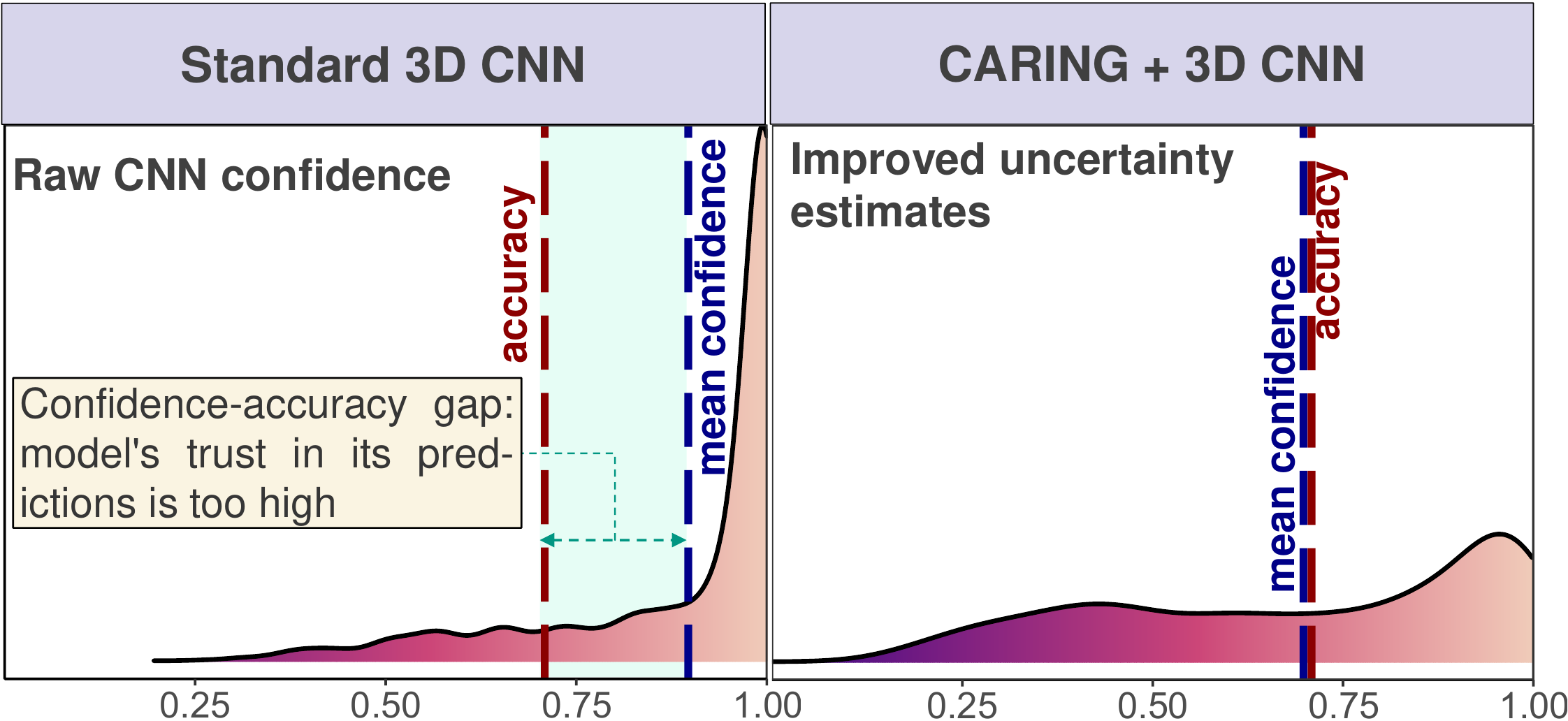}
    	 \captionof{figure}{\revised{An empirical example of experienced confidence-accuracy gap of modern video classification CNNs used for driver activity recognition. We visualize the distribution of the Softmax confidence values of the popular Pseudo 3D ResNet model deployed on the Drive\&Act dataset (validation split). The \textbf{X-axis} represents the space of possible probability estimates, while the \textbf{Y-axis} illustrates the frequency of samples with the corresponding probability estimate before (left) and after (right) the improvement through our model. Dashed red and blue lines depict the average confidence values and the accuracy accordingly which would match in a model with perfectly calibrated confidences.
		Native confidence values clearly underestimate model uncertainty (most samples received a confidence value of $>90\%$, while the accuracy is much lower). 	We therefore propose to  incorporate the \emph{\textbf{reliability}} of model confidences in the evaluation of driver observation models and introduce algorithms for improving it.} }
	\end{subfigure}
    \hspace{0.5cm}
    \begin{subfigure}[b]{0.4\linewidth}
	\includegraphics[width=\linewidth]{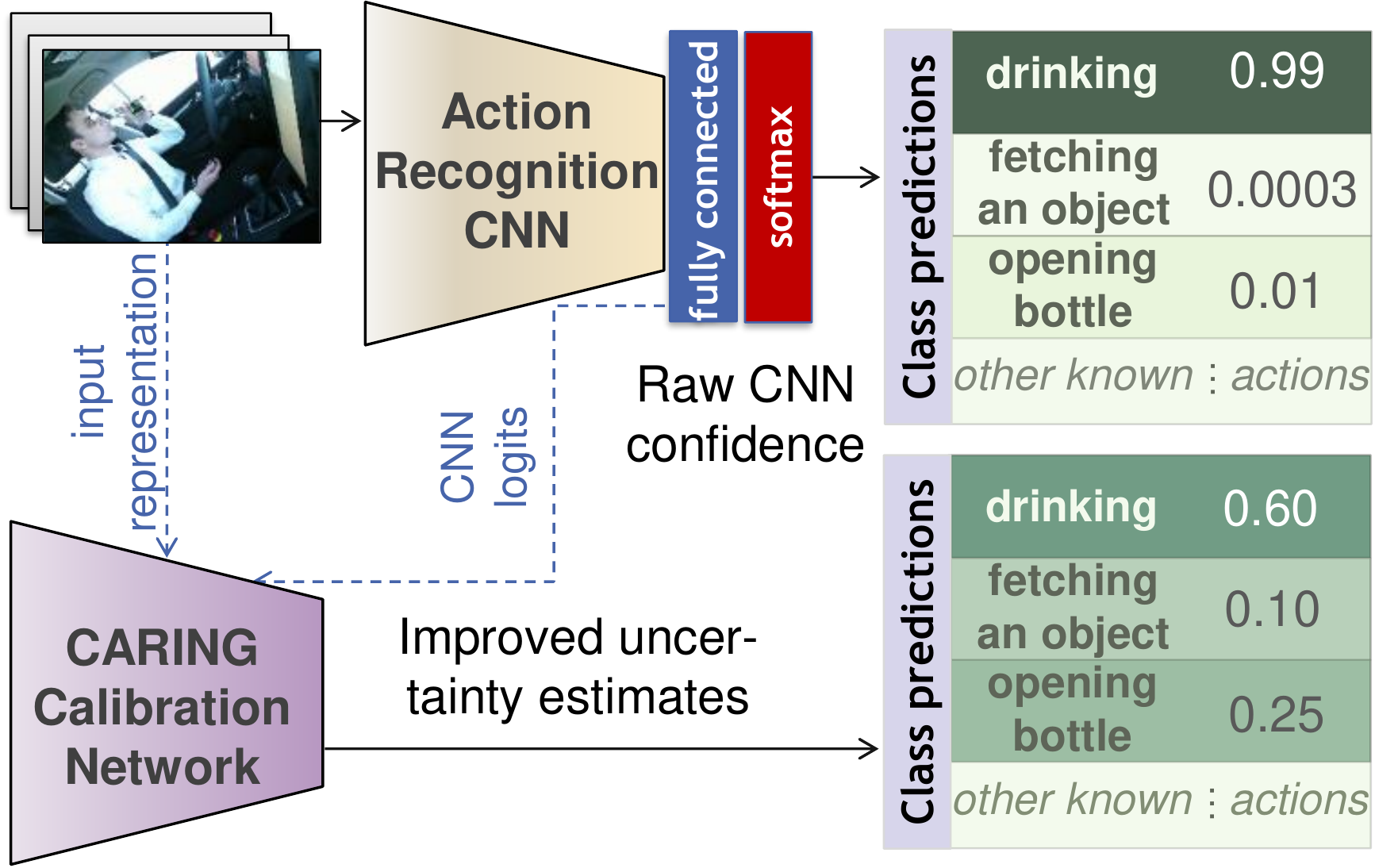}
	 \captionof{figure}{\revised{
	 An overview of the proposed \textbf{C}alibrated\textbf{ A}ction\textbf{ R}ecognition with \textbf{In}put \textbf{G}uidance (CARING) model for obtaining more realistic uncertainty estimates.
	CARING leverages an additional calibration neural network for learning to obtain realistic confidence estimates \textit{depending on the input}. The additional calibration network is optimized on a held-out validation set, during which the main classifier is frozen. It takes as input a video representation (intermediate features of the main classifier) and learns to infer an instance-dependent scaling factor, which is then used to scale the logits of the original network, therefore "softening" the resulting confidences and leading to much more realistic confidence estimates for driver observation. }}
	\end{subfigure}
	\hfill
	\caption{\revised{Observed problem of overconfident driver observation CNNs and an overview the proposed model for mitigating this effect.}}

		\label{fig:intro}
		\end{center}
	\end{figure*}
\begin{itemize}
	\item \textbf{Improved safety through identified distraction.} 
	Recent studies highlight that current activity directly affects  human cognitive workload in both, general- and driving   context~\cite{wolf2018estimating, deo2019looking, mok2015emergency}. For example, Deo and Trivedi~\cite{mok2015emergency} suggest that certain secondary activities such as \emph{interacting with the infotainment unit} negatively impact the perceived readiness-to-take-over.
	Therefore, the key application of such algorithms at SAE levels 0 to 3~\cite{sae2014automated} is the assessment of human distraction and reacting accordingly, for example, with a warning signal.
	
	\item \textbf{Increased comfort through automatic driver-centered adaptation.} 
	With the automation rising to SAE levels 4 and 5, increasing driver comfort by automatic adaptation of the vehicle controls becomes the more important use-case. 
	For example, movement dynamics might automatically adjust depending on the detected activity, \eg, softer driving if the person is drinking tea or sleeping. 
	
	\item \textbf{Novel intuitive communication interfaces.} 
	 Visually recognizing driver gestures might lead to novel communication interfaces inside the vehicle, serving  as a more intuitive alternative for the central console, as previous research identifies hand signals as a highly convenient way for human-machine interaction~\cite{gleeson2013gestures,  hand_gesture, roitberg2015multimodal}.

	\item \textbf{Identifying and preventing dangerous maneuvers. } 
		Since the majority of traffic fatalities are caused by inappropriate driving events induced by humans\cite{singh2015critical,lloyd2015reported},
		a further safety-related application of driver activity recognition during manual driving is intention prediction. 
	Timely anticipation of driver intention offers a possible solution to prevent potential accidents at an early stage, allowing ADAS to notice that the person \eg~is going to induce a dangerous turn and prevent the accident by taking over the control or notifying the driver~\cite{fang2021dada}.

\end{itemize}

\mypar{A Note on Terminology}
\revised{
While various meanings have been given to the term \textit{uncertainty}, we specifically refer to the ``classication uncertainty of a discriminative classification model'' as to \textit{inverse of model's confidence} ~\cite{gal2016uncertainty, malinin2019uncertainty}. 
We follow the probabilistic definition of \textit{model's confidence}~\cite{nixon2019measuring, dawid1982well, murphy1967verification, guo2017calibration}, which  is a numeric value depicting a probability estimate of a class membership in the context of multi-class categorization.
 Lastly, the \textit{reliability of model confidence} represents the quality of the observed confidence estimates, where we adopt the mathematical formulation of~\cite{guo2017calibration}, which intuitively means that if a network produces a confidence value of $95\%$ it should indeed be correct $95\%$ of times, (\ie this would also be the value of model's accuracy). The exact mathematical formulation of this definition is provided in Section \ref{sec:definition}.
 }

 \mypar{Contributions and Summary}
 This paper makes the first step towards  driver observation models  capable of identifying misclassifications by realistically estimating their prediction confidence and has following major contributions.
(1) We incorporate the Expected Calibration Error and the Brier Score metrics in the evaluation protocol of the large scale driver activity recognition dataset Drive\&Act~\cite{martin2019drive} and benchmark two CNN-based architectures commonly-used for driver observation in terms of their confidence reliability.
Our experiments indicate, that raw driver activity recognition models do not provide reliable uncertainty estimates and \textit{strongly overestimate their confidence}. To the best of our knowledge, this is the first study of the relation between the confidence of a driver activity recognition model and the likelihood of a correct prediction.
(2)
We augment driver activity recognition models with the temperature scaling approach~\cite{guo2017calibration} -- an off-the-shelf approach for model confidence calibration in image classification. 
This approach learns \emph{one global scalar}  $\tau$ through which the network logits are divided to obtain better confidence estimates.
(3) 
We expect, that taking the video input representation into account gives us cues which are helpful for estimating prediction uncertainty, and present a new method named \textbf{C}alibrated \textbf{A}ction \textbf{R}ecognition with \emph{\textbf{In}put \textbf{G}uidance} (CARING). 
A novel aspect of our approach is that CARING entails an additional calibration network, which learns to produce flexible temperature values $\mathcal{T}(\mathbf{z})$ \emph{dependent on the input} $\mathbf{z}$. In contrast,~\cite{guo2017calibration} uses one global scalar for scaling the confidences, so that the logits are always divided by the same value, while we dynamically infer the temperature values which suit currently observed state.

 Despite disregarding the input, temperature scaling alone leads to much more meaningful confidence values, while our input-dependent  CARING model excels these improvements, reducing the expected calibration error to only $\sim 5\%$ on Drive\&Act and to only $\sim 2.5\%$ in case of sufficient training data.
 However, our results also showcase room for improvement in case of underrepresented categories where the confidence-accuracy mismatch remains high even after significant improvement through CARING.

 This paper is an extension of our conference publication~\cite{roitberg2021uncertainty}, which has been expanded with specific focus on \textit{driver} behavior understanding, a detailed description of the proposed CARING method, and an extended set of experiments and analyses.

\section{Related Work}
\subsection{Driver Action Recognition}
Driver activity recognition is an essential task in Advanced Driving Assistance Systems (ADAS), aiming at recognition of normal- and abnormal driver behaviors, such as eating, phoning, and smoking.
According to~\cite{importance_driver_action_recognition}, over $90\%$ of vehicle crashes are due to drivers' abnormal behaviors and driving errors in the US, while the total number of traffic accidents might be decreased by $10-20\%$ if a good driving assistant monitoring system would be in place~\cite{deep_learning_driver_action_recognition}.
Similar to other computer vision fields, existing research works in driver action recognition field can be also divided into two main groups: (1) approaches based on manually defined features and (2) deep learning based approaches, which learn  intermediate representations end-to-end.

\mypar{Feature-based approaches.}
Until recently, the majority of driving observation frameworks comprised a manual feature extraction step followed by a classification module (for a thorough overview see~\cite{ohn2016looking}).
The constructed feature vectors are often derived from hand- and body-pose~\cite{hand_gesture, martin2018body,weyers2019action,li2019novel, hand_crafted_driver_action, hand_motion},
facial expressions and eye-based input~\cite{driver_gaze, eye_movement}, and head pose~\cite{head_pose_related, head_pose_estimation},
but also foot dynamics~\cite{foot_dynamics},  detected objects~\cite{weyers2019action}, and physiological signals~\cite{physical_signal} have been considered.
Classification approaches are fairly similar to the ones used in standard video classification, with LSTMs~\cite{jain2015car,martin2018body}, SVMs~\cite{ohn2014head,hand_gesture}, random forests~\cite{xu2014real} or HMMs~\cite{jain2015car}, and graph neural networks~\cite{li2019novel, martin2020dynamic} being popular choices.

\mypar{Deep learning-based approaches.}
Since the rise of deep learning, approaches based on Convolutional Neural Networks (CNNs), which operate directly on the image input and learn intermediate representations end-to-end, took over the top of most computer vision benchmarks, ranging from object classification and -detection to action recognition and semantic segmentation~\cite{he2016deep,wang2018nonlocal,qiu2017learning, carreira2017quo, feng2020deep,peng2021mass,ma2021densepass}.
These developments also had strong influences on the field of driver observation where a variety of works report top performance of CNN-based architectures~\cite{kouchak2020detecting,chen2020driver,reiss2020activity,qin2021distracted, martin2019drive, deep_learning_driver_action_recognition, Roitberg2020_InterpretableCNN, wharton2021coarse,ortega2020dmd,reiss2020deep, abouelnaga2018real}, with spatiotemporal CNNs, such as I3D~\cite{carreira2017quo}, P3D~\cite{qiu2017learning}, and 3D-MobileNet~\cite{kopuklu2019resource} being popular backbones~\cite{ortega2020dmd,martin2019drive,reiss2020deep,Roitberg2020_InterpretableCNN,kopuklu2020drivermhg}.
Further accuracy improvements are often achieved through techniques such as computation and fusion of additional modalities (\eg, semantic segmentation, optical flow)~\cite{deep_learning_driver_action_recognition, gebert2019end} or use of attention modules~\cite{importance_driver_action_recognition, wharton2021coarse} and CNN-RNN combinations~\cite{gebert2019end, driver_recurrent,cognitive_prior}.

The general goal of the above research is to improve the recognition accuracy on driver activity recognition datasets without considering whether the network \emph{confidence indeed reflects the likelihood of a correct prediction} (Figure~\ref{fig:caring-model} demonstrates the confidence-accuracy mismatch of P3D on the Drive\&Act~\cite{martin2019drive} dataset revealed by our study).

In our work, we go beyond the incentive of high top-1 accuracy and aim for driver activity recognition models with \textit{realistic confidence estimates}. 
Keep in mind, that while our method leads to far more realistic uncertainty estimates, the accuracy is not impacted, as the transformation scales the  output without changing the category order.

\subsection{Detecting  Model Misclassifications}

Since real applications of deep CNNs need good quality of confidence measures, several researchers have highlighted this requirement~\cite{sunderhauf2018limits,hendrycks17baseline,nguyen2015deep}.
However, this issue still lacks attention in both, driver- and general activity recognition.
Still, multiple works have addressed this topic in classical machine learning~\cite{niculescu2005predicting,degroot1983comparison,platt1999probabilistic}, image classification~\cite{guo2017calibration,gal2016dropout, hendrycks17baseline}, and person identification~\cite{bansal2014towards}, offering useful insights for applying these techniques for video-based driver observation.
A group of those methods follows the Bayesian approach, such as Monte Carlo Dropout sampling~\cite{gal2016dropout} or using an ensemble of models to estimate uncertainty~\cite{lakshminarayanan2017simple}.
Compared to dropout sampling, which gives gaussian distributions to assess model uncertainty,  calibration-based techniques~\cite{platt1999probabilistic, zadrozny2001obtaining,zadrozny2002transforming,guo2017calibration,ott2018analyzing,naeini2015obtaining} return a single confidence value. 
Calibration-based methods hold several advantages: they do not require sampling and are therefore less computationally expensive and their result is \textit{calibrated}, meaning that the resulting  values ideally match human interpretation of probabilities (\ie~the likelihood of a correct outcome).
Guo~\etal\cite{guo2017calibration} compared several calibration-based approaches such as isotonic regression, histogram binning, and Bayesian quantile binning regression on image and document classification tasks.
In their work, they introduced  \textit{temperature scaling} -- an approach, which is a variant of the Platt Scaling approach~\cite{platt1999probabilistic}, in which a single global parameter is learned on  a validation set, and then utilized to scale the network logits and make the output confidence closer to the actual accuracy.
Temperature scaling was simple but proved to be highly effective,  outperforming far heavier approaches~\cite{guo2017calibration} and later found success in natural language processing~\cite{ott2018analyzing} and medical applications~\cite{huang2020tutorial}.

We for the first time integrate different building blocks for identifying the faithfulness of model confidences  for driver activity analysis.
In particular, we augment driver activity recognition networks with  uncertainty estimation through (1) the raw Softmax confidences~\cite{hendrycks17baseline},  (2) the temperature scaling technique of Guo \etal~\cite{guo2017calibration}, and (3) our newly proposed CARING model, which, in contrast to~\cite{guo2017calibration}, features \textit{input-driven} confidence transformation. 
While in  temperature scaling, the logits are always divided by the same global value, our CARING model dynamically infers different scaling parameters for different inputs through an additional calibration network.

\section{Uncertainty-sensitive Action Recognition}

We aim to explore \textit{reliability of model confidence} in the field of driver observation, where the existing frameworks have been mostly motivated by high accuracy~\cite{hand_gesture, martin2019drive, abouelnaga2018real, deep_learning_driver_action_recognition}. 
First, we \revised{formalize} the problem and equip the \emph{Drive$\&$Act} benchmark~\cite{martin2019drive} with multiple metrics linked to the confidence quality (Section ~\ref{sec:definition}). 
We then combine two 3D CNNs (model description in Section \ref{sec:backbone}) often used for driver activity recognition, with three different strategies for quantifying prediction confidence: using raw \textit{Softmax} output (Section \ref{sec:softmax}), transforming the confidences through temperature scaling (Section \ref{sec:temperature}) and our newly proposed CARING model, which learns to infer optimal scaling parameters depending on the input (Section \ref{sec:caring}).

\subsection{Problem Definition: Reliable Confidence Measures}
\label{sec:definition}

Intuitively, the confidence of a classification model should reflect its accuracy, \ie,  \emph{reliable} confidence should indicate the probability of a prediction to be correct.
Based on a driver observation $x$, we predict an action class $a_{pred}$ with model confidence $\hat{p}(a_{pred}) \in [0,1]$, $a \in A_{\{1,...,m\}}$ being the set of possible driver behaviour classes. 
Provided with the ground-truth label $a_{true}$, we consider our model confidence to be well calibrated and reliable if it, on average, reflects  the probability of a successful outcome $\mathbb{P}(a_{pred}= a_{true})$. 
We follow the definition of~\cite{guo2017calibration} and  assume that the confidence quality of a model is perfect if the following condition is met:
\begin{equation}
    \mathbb{P} (a_{pred}= a_{true}| \hat{p} (a_{pred}) = p) = p, \quad \forall p \in [0, 1]
    \label{eq:perfect_calibration}
\end{equation}

The model therefore delivers  reliable probability estimates if the mismatch between the average model's confidence and the prediction accuracy is low (the right side of \figref{fig:intro} illustrates such confidence-accuracy disarray).

In practice,  perfect model calibration and even its \textit{perfect} evaluation are not possible, since Equation~(\ref{eq:perfect_calibration}) is defined on the continuous interval $[0,1]$ while we only have a finite amount of confidence estimates $\hat{p}(a_{pred})$. 
However, this value can be  \textit{approximated} with different techniques, \eg, discrete partitioning of the probability space. 
We consider three different metrics: Expected Calibration Error (ECE)~\cite{guo2017calibration}, the presumably most popular metric for quantifying confidence discrepancy, the normalized Brier score~\cite{brier1950verification}, and Negative Log Likelihood. 

\begin{figure}\centering
	\includegraphics[width=0.5\textwidth,trim={1.8cm 4.6cm 2.25cm 3.0cm},clip]{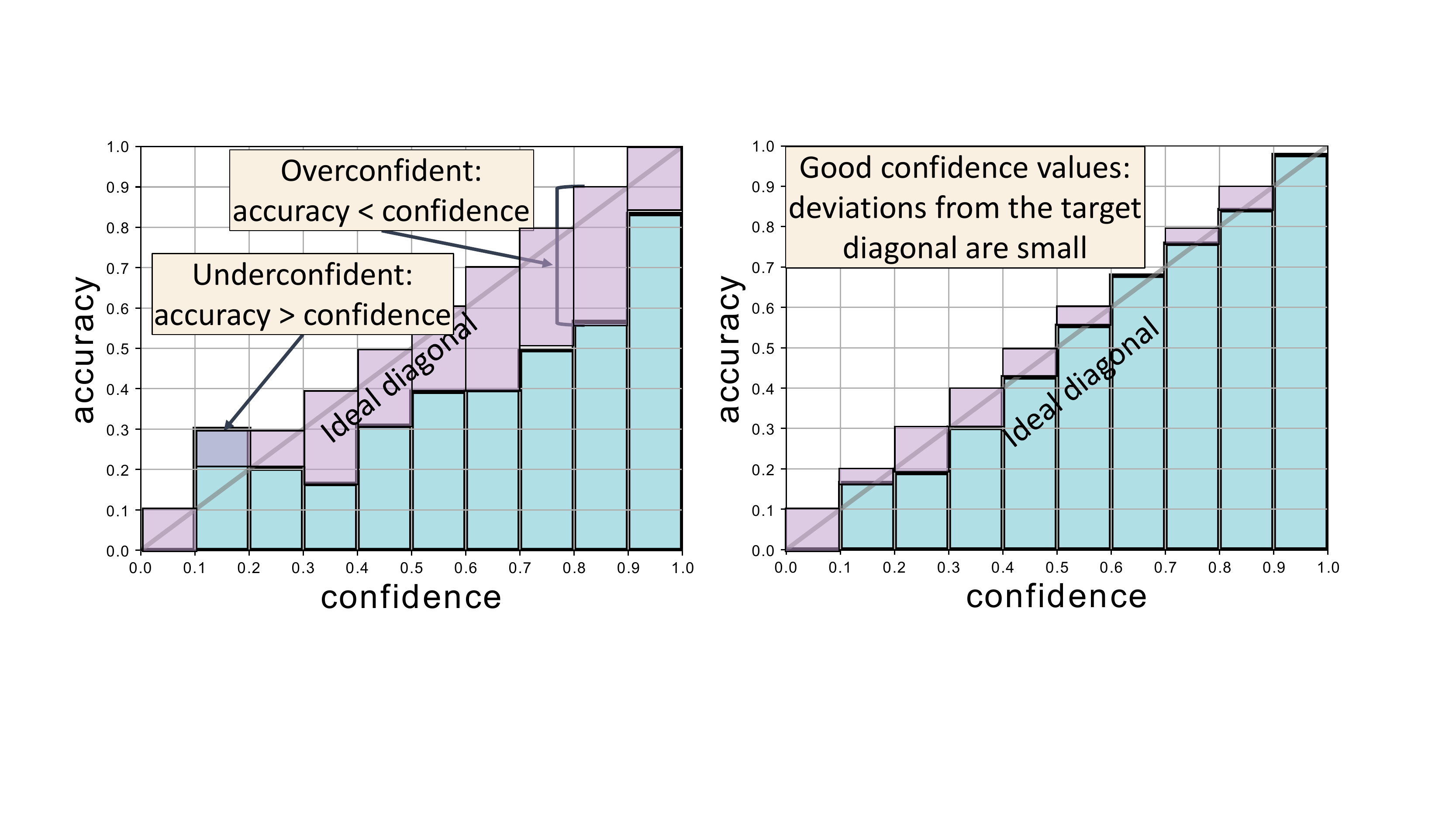}
	\caption{Reliability diagrams of a model with poor confidence estimates (left) and a well-calibrated model (right). The illustrated data are the confidence values of the Pseudo 3D ResNet a the Drive\&Act validation split before and after the improvement with the CARING calibration network.
	}
	\label{fig:reliability_explanation}
	\vspace{-0.3cm}
\end{figure}

\mypar{Expected Calibration Error (ECE).}
One way to solve the issue of continuous probability space in Equation (\ref{eq:perfect_calibration}) is quantizing the $[0,1]$ interval into equally sized segments $seg_i = \left [\frac{i}{K}, \frac{i + 1}{K} \right)$ with $i \in {0,...,K-1}$ ($K=10$ in this work) and calculating the average accuracy and  confidence among all $N_{seg_i}$ samples falling into the same confidence segment (see Figure~\ref{fig:reliability_explanation}). 
In case of perfect confidence calibration, the  accuracy-confidence distance of the individual segments should be zero. 
The Expected Calibration Error (ECE)~\cite{guo2017calibration}, leverages such partitioning and measures the absolute difference between accuracy $acc(\mathrm{seg_i})$ and average confidence $\hat{p} (seg_i)$ for each individual segment $seg_i$, followed by averaging among all segments weighted by the number of samples in each of the $K$ segments. 
Formally, ECE is computed as:

\begin{equation}
ECE = \sum_{i = 1}^{K} \frac{N_{seg_i}}{N_{total}} |acc(seg_i) - \hat{p} (seg_i)|,
\label{eq:ece}
\end{equation}
where $N_{total}$ denotes the number of samples in all segments.
In deep learning research, ECE is perhaps  the most wide-spread measure of model confidence reliability~\cite{guo2017calibration, ovadia2019can, huang2020tutorial, naeini2015obtaining}, and we therefore also view it as our primary performance metric.

\mypar{Reliability Diagrams.} ECE can be effectively visualized using \emph{reliability  diagrams} (Figure \ref{fig:reliability_explanation}), where the X-axis represents the probability space discretized into $K$ bins.
In each bin, we visualize a bar with size matching the empirically computed accuracy of samples, which confidence fell into this segment. 
Since the accuracy and the confidence match in perfect case, the bars should ideally match the diagonal. 
Our results in Figure~\ref{fig:reliability_explanation} showcase, that for an out-of-the-box (illustrated on the left), the accuracies are significantly \emph{below} the confidence value. The original model is therefore \emph{overly confident}.

\mypar{Brier Score.}
Another metric for quantifying model confidence quality 
is the Brier score~\cite{brier1950verification}, initially developed to quantify weather forecasts reliability and later adapted as a general confidence reliability measure. 
The Brier score represents squared error between the estimated probabilities and the  one-hot encoding of the ground-truth data. 
As the original definition has different boundaries for binary and multi-class cases~\cite{brier1950verification}, we use the normalized version of the Brier score, where the values lie between 0 and 1, formally defined as~\cite{kruppa2014probability}: 

\begin{equation}
Brier = \frac{1}{2N_{total}} \sum_{i = 1}^{N_{total}} \sum_{a = 1}^{m} (C_{a,i} - \hat{p}(a, i))^2 ,
\end{equation}
where $N_{total}$ is the total number of samples, $m$ is the number of possible driver activity classes, $\hat{p}(a, i)$ is the model probability estimates of the $i$th sample belonging to category $a$ and  $C_{a,i} \in {0,1}$ is $1$ if $a$ is the ground-truth class of the $i$th sample, otherwise 0.
Note, that previously published research describes Brier score as a metric is \textit{insensitive towards rare classes} in the test set~\cite{ovadia2019can} and  should be taken with caution since in real-life datasets the categories are rarely equally distributed (which is also the case in \textit{Drive\&Act}).

\mypar{Negative Log Likelihood (NLL).}
Negative Log Likelihood,  which is also leveraged for optimizing the classifier itself, can be used to measure quality of the uncertainty estimates~\cite{guo2017calibration} and is therefore reported in our study.

\subsection{Backbone Neural Architectures}
\label{sec:backbone}

We evaluate model confidence reliability on two widely used architectures, the Pseudo 3D ResNet~\cite{qiu2017learning} as well as the Inflated 3D ConvNet~\cite{carreira2017quo}.
Both networks have proven to be  successful in driver observation~\cite{martin2019drive,ortega2020dmd}.
While the P3D ResNet decouples spatial and temporal convolutions by either using a ($3\times3\times1$) filter kernel for the spatial domain or a ($1\times1\times3$) filter kernel for the temporal domain, the I3D Network makes use of filter kernels which perform spatial and temporal convolutions at the same time.
Another structural difference, is that P3D ResNet has an impressive depth of $152$~layers enabled through use of residual connections to ease the gradient flow, while I3D has $27$~layers and stacks multiple Inception modules, which perform different types of convolutions in parallel.

\subsection{Softmax Output as Confidence Measures}
\label{sec:softmax}

Similar to other neural networks, the last fully-connected layer are referred to as a \emph{logit vector} $\textbf{y}$.
The individual values $y_a$ are scores marking how likely the activity $a$ is indeed the current class.
The raw \emph{logit vector} is \emph{not normalized}.
In order to mimic a probability function (have values between $0$ and $1$  and sum up to $1$), it is commonly normalized using \textit{Softmax}.
Although these scores are rather \emph{pseudo-probabilities}, they are often used as model uncertainty estimates~\cite{hendrycks17baseline}.
The confidence estimates in this case are obtained as:

\begin{equation}
    \hat{p} (a_{pred}) = \max\limits_{a \in \mathcal{A}} \frac{exp(y_a)}{\sum\limits_{\tilde{a} \in \mathcal{A}} exp(y_{\tilde{a}})}
\end{equation}

\subsection{Calibration via Temperature Scaling}
\label{sec:temperature}

\begin{figure*}[!t]
	\centering
	\includegraphics[width=\linewidth]{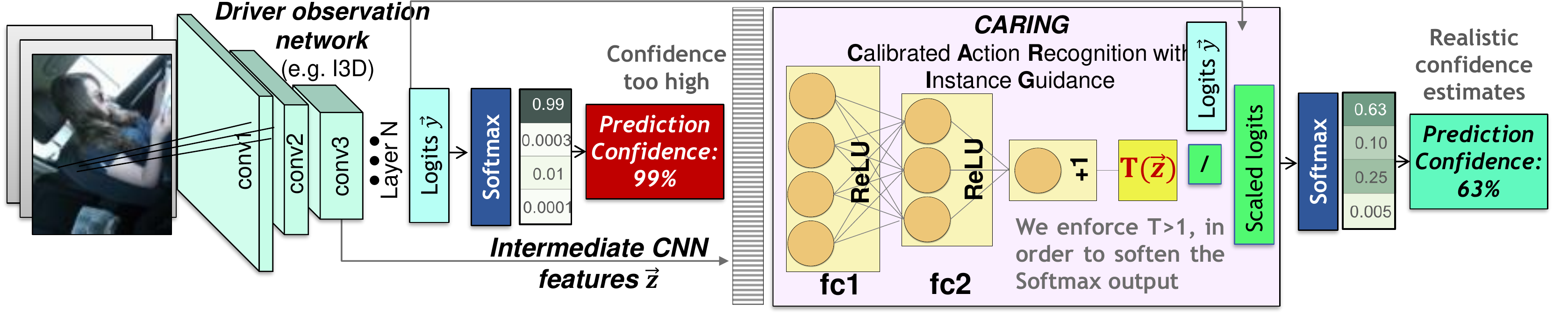}
	\caption{Overview of the Calibrated Action Recognition under Instance Guidance Model (CARING). CARING is an additional neural network which learns to infer the scaling factor $\mathcal{T}$ depending on the instance representation. The logits of the original activity recognition network are then divided by $T$, giving better estimates of the model uncertainty. \revised{In our implementation, the intermediate vector $\textbf{z}$ is extracted from the 9th inception block of the I3D model, while for P3D we use the layer before the last fully-connected layer.}}
	\label{fig:caring-model}
		\vspace{-0.3cm}
\end{figure*}

In image recognition, an off-the-shelf strategy for improving confidence estimates is \emph{temperature scaling}~\cite{guo2017calibration}.
Temperature scaling leverages \textit{a single, global parameter $\tau$} to scale the logit vector prior to \textit{Softmax} normalization, with the final estimates becoming:

\begin{equation}
    \hat{p} (a_{pred}) = \max\limits_{a \in \mathcal{A}} \frac{exp(y_a / \tau)}{\sum\limits_{\tilde{a} \in \mathcal{A}} exp(y_{\tilde{a}} / \tau)}
\end{equation}

This introduces the properties $\lim\limits_{\tau \to \infty}\hat{p} (a_{pred}) = 1/m$ and $\lim\limits_{\tau \to 0}\hat{p} (a_{pred}) = 1$, meaning that $\tau$ is a parameter which either decreases confidence ($\tau > 1$) or increases confidence ($\tau < 1$).
After the classifier is trained to assign the categories in a standad way, the model weights are frozen and the parameter $\tau$ is optimized on a held-out validation set also using Negative-Log-Likelihood. 
\revised{Note, that this scaling affects probability estimates of all classes equally and does not alter the order of the top-k categories or the final top-1 prediction of the model. This means that the accuracy of the model stays the same.}
Although the method is remarkably simple, it had been extremely effective for obtaining well-calibrated image recognition CNNs, surpassing more complex approaches~\cite{guo2017calibration}.
We learn a suitable temperature parameter $\tau$ using Gradient Descent with a learning rate of $0.01$ for $50$ epochs.

\subsection{Calibrated Action Recognition with Input Guidance}
\label{sec:caring}

Next, we introduce a new approach for obtaining good uncertainty estimates by learning to scale the logits \emph{depending on the input}.
While in the next section our experiments will reveal, that temperature scaling significantly improves the confidence reliability on its own, it \textit{does not consider the current observation} when transforming the confidences, as the logits are always divided by the same temperature scalar $\tau$.
Our idea is leveraging information present in the current driver observation, as it may carry important hints about the prediction reliability (\ie~the sample might contain visible noise).
Therefore, we propose to infer an individual input-dependent scaling factor $\tau(z)$ during inference instead of learning a single global parameter $\tau$, resulting in the confidence calculation.
Our model and the temperature scaling approach~\cite{guo2017calibration} therefore have one crucial difference: the scaling factor is not fixed but varying dependent on current input.
We therefore  dynamically infer the scaling parameter $\mathcal{T}(\textbf{z})$ on-the-fly at test-time depending on the video representation $\textbf{z}$. Therefore, the transformed logits become $\textbf{y}_{scaled} = \textbf{y}/\mathcal{T}(\textbf{z})$.

To enable such input-dependent inference of $\mathcal{T}(\textbf{z})$, we propose an additional \emph{calibration neural network}, which we refer to as  \textit{CARING}  (\textbf{C}alibrated \textbf{A}ction \textbf{R}ecognition under \textbf{In}put \textbf{G}uidance), as it guides the logit transformation depending on the video input.
An overview of our model is provided in Figure \ref{fig:caring-model}.
CARING is a lightweight regression network with two fully-connected layers and the output of the second layer being a single regressed value.    
The final scaling factor is obtained by applying the \textit{relu} activation and adding $1$ to assure $\mathcal{T}(\textbf{z})\geq1$, so that the probability estimates are softened.
The input-dependent temperature $\mathcal{T}(\textbf{z})$ is therefore obtained as:
\begin{equation}
\quad \mathcal{T}(\mathbf{z}) = 1 +relu( \mathbf{W_2}~relu( \mathbf{W_1} \mathbf{z} + \mathbf{b}_1)  + \mathbf{b_2}), 
\end{equation}
where $\mathbf{W_1}$,$\mathbf{W_2}$ depict the weight matrices, and  $\mathbf{b_1}$,  $\mathbf{b_2}$ are the bias terms, and $\mathbf{z}$ is an intermediate representation of current input.
We leverage the intermediate feature vector of the backbone classification network, which have dimensionality of $1024$ and  $2048$  for Inflated 3D ConvNet and Pseudo 3D ResNet, respectively.

After we regress the input-dependent temperature $\mathcal{T}(\mathbf{z})$, the logits are divided by it before the \emph{Softmax} function and our final confidence estimates become:
\begin{equation}
\hat{p} (a_{pred}) = \max\limits_{a \in \mathcal{A}} \frac{exp(\frac{y_a}{\mathcal{T}(\mathbf{z})})}{\sum\limits_{\tilde{a} \in \mathcal{A}} exp(\frac{y_{\tilde{a}}}{\mathcal{T}(\mathbf{z})})}.
\end{equation}

\mypar{Training procedure.}
\revised{
Conceptually, our proposed architecture comprises two main components: 1) the standard classification network aimed at assigning the correct driver behaviour to the input video and 2) an additional calibration network used to scale the classifier logits and produce more realistic confidences. 
First, we train the classifier following the standard procedure and Negative-Log-Likelihood (NLL) optimization.
In the second step, the classifier weights are frozen and the calibration network is optimized on a separate, held-out validation set which examples were not present during the classifier training. 
Different losses can be used for training such network, including the NLL loss~\cite{guo2017calibration}, Maximum Mean Calibration Error (MMCE)~\cite{kumar2018trainable}, Brier Loss~\cite{brier1950verification} and Focal Loss~\cite{mukhoti2020calibrating}. We selected the NLL loss in our training, since it is presumably the most popular choice~\cite{guo2017calibration,ott2018analyzing,huang2020tutorial} for confidence calibration and is also used in the temperature scaling approach of~\cite{guo2017calibration} to which our model is compared. }

\revised{
 By freezing the model weights and only training our lightweight calibration model on a validation set, a realistic distribution of correct and incorrect predictions is provided and being overly confident about these predictions would increase the loss. 
 In the same fashion as temperature scaling, CARING is a post-processing technique for improving model probability estimates.  
It \emph{does not influence the accuracy}, as the order of the logit categories does not change (we do divide by different temperature values for different observations, but all the logit vector values are divided by the same scalar).
 We optimize CARING with a learning rate of $5\mathrm{e}{-3}$ and a weight decay of $1\mathrm{e}{-6}$  for 300 epochs on a held-out validation set using the previously mentioned NLL loss while maintaining the weights of the backbone classification network frozen.
The complete process is described as pseudo-code in Algorithm ~\ref{alg:training}.
}

\begin{minipage}[ct]{0.95\columnwidth}
\begin{lstlisting}[
    % language=python,
   % caption={The CARING training and evaluation %procedure in the specific case of using I3D as a %backbone and evaluating on the ECE loss.}, 
    label={code:training}, 
    captionpos=b, 
    breaklines=true,
    escapechar=\%,
    basicstyle=\fontsize{8}{9}\selectfont\ttfamily\color{black}
]

%\textit{\textbf{Training of I3D}}%
train, valid, test = split(Drive & Act)

for epoch in [1...300]:
  for data, labels in train:
	logits = i3d(data)
	loss = NLL(logits, labels)
    backpropagate(loss)

%\textit{\textbf{Training of CARING}}%
caring = Sequential(
  Linear, ReLU, 
  Linear, ReLU,
  Lambda(x -> x + 1))
  
for epoch in [1...300]:
  for data, labels in valid:
    logits = i3d(data)
    intermediate_features = truncate(i3d)(data)
    temperature = caring(intermediate_features)
    scaled_logits = logits / temperature
    loss = NLL(scaled_logits, labels)
    backpropagate(loss)

%\textit{\textbf{Inference and evaluation with ECE}}%
for data, labels in test:
  logits = i3d(data)
  intermediate_features = truncate(i3d)(data)
  temperature = caring(intermediate_features)
  scaled_logits = logits / temperature
  logits_i3d.append(logits)
  logits_caring.append(scaled_logits)
  labels_all.append(labels)

%\textit{\textbf{(See Eq. \ref{eq:ece} for ECE computation)}}%
ECE_i3d = ECE(logits_i3d, labels_all)
ECE_caring = ECE(logits_caring, labels_all)

\end{lstlisting}
\captionof{algorithm}{\revised{The CARING training and evaluation procedure in the specific case of using I3D as a backbone and evaluation using the Expected Calibration Error (ECE) metric.}}
\label{alg:training}
\vspace{0.5cm}
\end{minipage}

\revised{For the intermediate representation vector $\textbf{z}$, we leverage the output of the 9th inception block of I3D.
We  average the output over the temporal dimension to reduce the vector size, resulting in $1024$-dimensional vectors. 
For P3D, we use the layer before the last fully-connected layer (the output size is $2048$). 
Furthermore, we observe that the CARING performance does not alternate significantly  if the representations were extracted at medium or late network depths, but the quality declines at earlier layers. 
Note, that the size of $\textbf{z}$ directly impacts computational complexity of CARING. 
For our configuration, (\textit{i.e.}, 1024-dimensional $\textbf{z}$ in case of I3D), CARING comprises 66.7 parameters and requires 0.13 MFLOPS for a forward pass, which is a small overhead given the significantly larger size of the classification network (12.06 M parameters and 107.9 GFLOPS for I3D).
}

\emph{\revised{Why is training the calibration network with NLL loss effective against miscalibration?}}
\revised{Since both the classifier backbone and the CARING model use the NLL loss, we need to clarify why such optimization leads to  unrealistic confidences after the standard classifier training but gives proper confidence estimates after the calibration. 
NLL itself indeed  reflects confidence miscalibration~\cite{guo2017calibration} and we therefore use it as one of our  metrics.}
The main reason for the confidence-accuracy disarray rising during the classifier training is \textit{overfitting on the training set combined with the target labels being exclusively  0 or 1}~\cite{guo2017calibration,kumar2018trainable}.
Target labels for classification training are commonly 0 or 1 \revised{(one-hot encodings of the ground truth)}.
\revised{The high capacity of modern neural architectures leads to overfitting to the one-hot encoded labels, which is considered the key reason for model miscalibration~\cite{guo2017calibration, mukhoti2020calibrating} (although other factors have been witnessed, such as batch normalization). }
By training a model to learn \textit{Softmax} scores which mimic binary targets, it is incited to provide overly confident results. 
\revised{After the classifier training has converged in terms of \textit{accuracy} (which is usually very high on the training data), it further optimizes the NLL-criteria to match the $0/1$ labels, which will usually be $1$ for the predicted class on the training data, leading to overly confident models. 
Consequently, the classifier becomes overly confident due to the unrealistically high amount of the correct $1$-label predictions.}
By freezing the model weights  and only training our lightweight CARING model on a held-out set, \revised{we bypass the issue of overfiting to 0/1 labels due to high model capacity}. A realistic distribution of correct and incorrect predictions is therefore provided and being overly confident about these predictions would increase the loss. Note that overfitting is prevented since CARING does not influence the model accuracy (\revised{since the classifier is frozen and the logits order stays the same}).

\begin{figure}
	\centering
	\begin{subfigure}{.25\textwidth}
		\centering
		\includegraphics[trim={0.4cm 0 0 0},clip,width=\linewidth]{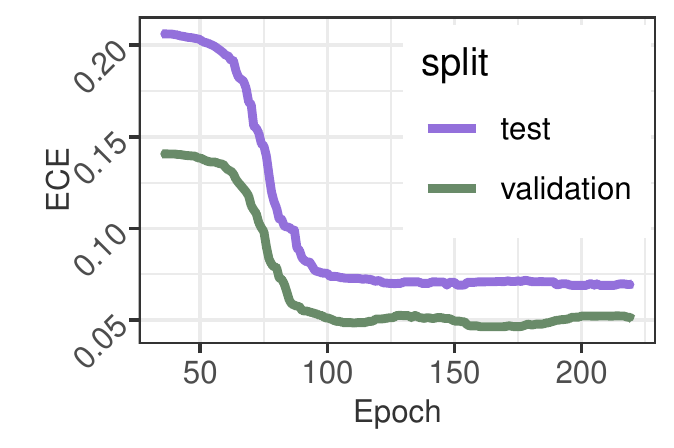}
		\caption{Expected Calibration Error improvement during the training procedure for validation  and test data.}
		\label{fig:ece_during_training}
	\end{subfigure}%
	\begin{subfigure}{.25\textwidth}
		\centering
		\includegraphics[trim={0.4cm 0 0 0},clip,width=\linewidth]{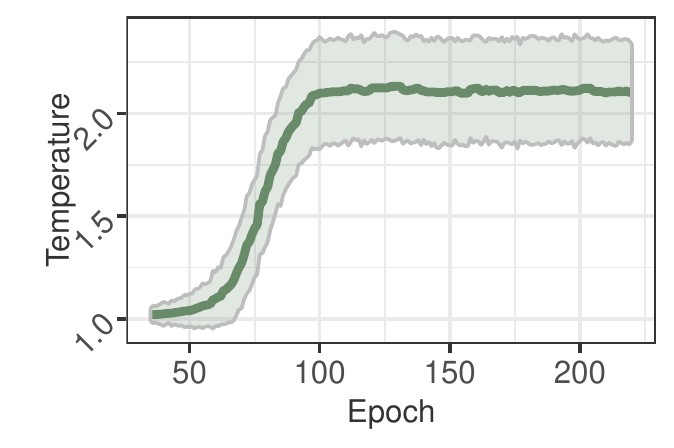}
		
		\caption{Average temperature and its\\ \null \quad standard deviation estimated by \\ \null \quad our model during training.}
		\label{fig:avg_temperature_during_training}
	\end{subfigure}
	\caption{CARING model evolution during training for one Drive\&Act split. Both average value and standard deviation of the learned input-dependent scaling parameter  $\mathcal{T}\vec(z)$ rise as the training proceeds (right figure). Jointly with the decrease of the calibration error (left figure), this indicates the usefulness of learning different scaling parameters for different inputs.}
	\label{fig:learning_procedure}
		\vspace{-0.3cm}
	
\end{figure}

\textit{Does our \textit{CARING} model indeed scale different inputs differently?}
Although CARING always takes in representation of the video input, it could theoretically ignore it and fall back learning a constant value, therefore converging to  conventional temperature scaling.
In order to validate if our model actually infers different   $\tau(z)$ for different inputs, we analyze the Expected Calibration Error (Figure~\ref{fig:ece_during_training}) in combination with the  mean and standard deviation of the scaling factor $\tau(z)$ (Figure~\ref{fig:avg_temperature_during_training}) computed over all validation samples during each CARING training epoch.
Figure~\ref{fig:learning_procedure} reveals that while the ECE decreases during optimization, the standard deviation of the individual scaling factors increases, indicating that varying $\tau(z)$ is indeed beneficial for confidence reliability, which will also be validated empirically in the upcoming section.

\section{Experiments}

\revised{
In this section, we for the first time benchmark modern driver activity recognition models with regard to their abilitiy to identify failure cases through their own probability estimates.
To this intent we for the first time consider metrics described in Section in the context of driver activity recognition ~\ref{sec:definition}.
In a variety of settings featuring different input modalities and amount of training data, we benchmark two prominent video classification CNNs, against their versions equipped with the temperature scaling method~\cite{guo2017calibration} and our proposed CARING approach.
We hope that by establishing this benchmark we will motivate researchers to rethink the
roles of uncertainty estimates in driver observation.
} 

\revised{
This Section is structured as follows. First, we give an overview of the evaluation testbed and the underlying data (Section \ref{sec:benchmark-settings}).
Next, we provide a thorough  evaluation of confidence estimates for driver activity recognition, focusing on the main evaluation setting of Drive\&Act (a single NIR modality) and distinguishing between different amount of available training data (Section \ref{sec:exp-closed-set}).
Since different activity types might be relevant for different driving applications, Section \ref{sec:exp-different} is devoted to individual category-specific outcomes. 
Then, in Section \ref{sec:exp-views} we focus on recognition from different sensor types and placements within the vehicle cabin, demonstrating generalization capabilities of our model.
Finally, Section \ref{sec:calib_diag} provides qualitative analyses of the resulting distributions of model confidences.
}

\subsection{Benchmark settings}
\label{sec:benchmark-settings}

Currently, there is no established evaluation scheme targeting the reliability of confidence values in the context of driver observation. 
Thereby, we adapt the  evaluation protocols of Drive\&Act~\cite{martin2019drive} -- a large-scale driver activity understanding testbed, to our task.
Drive\&Act~\cite{martin2019drive}, is a multimodal benchmark for driver activity recognition covering nine synchronized views and three modalities (near-infrared, depth, and color). Drive\&Act comprises $34$ fine-grained activity classes on the main evaluation level. Yet, the categories are heavily unbalanced, which is typical for application-specific datasets: the number of samples ranges from only $19$ examples of \emph{taking laptop from backpack} to $2797$ instances of \emph{sitting still}.
Considering that CNNs often suffer when learning from few examples, the behaviors are sorted according to their frequency in the dataset and separated into \emph{common} (top half of the classes) and \emph{rare} (the bottom half) ones.
The models are subsequently evaluated in three different modes: incorporating \emph{all activities} as in the usual case, using only the \emph{overrepresented-}, or only the \emph{rare} categories.
\looseness=-1

\begin{table}[!t]

	\scalebox{0.85}{
	\begin{tabular}{ @{}lcccccc}
		\toprule
		\multirow{2}{*}{\textbf{Model}} & \multicolumn{2}{c}{\textbf{ECE (\%)}} &
		\multicolumn{2}{c}{\textbf{Brier}} &\multicolumn{2}{c}{\textbf{NLL}} \\
		& val & test & val & test &val & test  \\ 
		\midrule
		\textbf{Drive\&Act - Common Classes} \\
		P3D \cite{qiu2017learning} \scriptsize{\textcolor{table_standard}{\textbf{\circled{S}}}}&16.9 & 19.39 & 0.27&	0.32& 1.63 & 1.85  \\
		I3D \cite{carreira2017quo} \scriptsize{\textcolor{table_standard}{\textbf{\circled{S}}}}&10.22 & 13.38 & 0.12	& 0.16  & 0.90 & 1.27 \\
		P3D + Temperature Scaling  \cite{guo2017calibration}\scriptsize{\textcolor{table_uncertainty}{\textbf{\circled{U}}}}&5.65 & 5.70 &0.25&	0.29& 1.28 & 1.48\\ 
		I3D + Temperature Scaling \cite{guo2017calibration}\scriptsize{\textcolor{table_uncertainty}{\textbf{\circled{U}}}}&5.31 & 6.99 &0.11 &	\textbf{0.15} & 0.57 & 0.83  \\
		CARING - P3D (ours) \scriptsize{\textcolor{table_uncertainty}{\textbf{\circled{U}}}} &4.81 & \textbf{4.27} & 0.24&	0.28&  1.19 & 1.42\\
		CARING - I3D (ours) \scriptsize{\textcolor{table_uncertainty}{\textbf{\circled{U}}}}&\textbf{2.57 }& 5.26 & \textbf{0.10} &	\textbf{0.15}&  \textbf{0.50} & \textbf{0.78}  \\
		\midrule
		\textbf{Drive\&Act - Rare Classes} &  &  & &  \\
		P3D \cite{qiu2017learning} \scriptsize{\textcolor{table_standard}{\textbf{\circled{S}}}}&31.49 & 37.25 &  0.39&	0.50& 3.43 & 4.68 \\
		I3D \cite{carreira2017quo} \scriptsize{\textcolor{table_standard}{\textbf{\circled{S}}}}&31.48 & 43.32  &0.34 &	0.47 & 3.41 & 4.54\\
		P3D + Temperature Scaling \cite{guo2017calibration} \scriptsize{\textcolor{table_uncertainty}{\textbf{\circled{U}}}}&17.83 & 21.09 & 0.34&	0.42&   2.26 & 2.99\\
		I3D + Temperature Scaling \cite{guo2017calibration} \scriptsize{\textcolor{table_uncertainty}{\textbf{\circled{U}}}}&24.97 & 32.38 & 0.30	& 0.41&  1.96 & 2.62\\
		CARING - P3D  (ours) \scriptsize{\textcolor{table_uncertainty}{\textbf{\circled{U}}}}&\textbf{13.73 }& \textbf{19.92} &0.33&	0.42& 2.12 & 2.93\\
		CARING - I3D (ours) \scriptsize{\textcolor{table_uncertainty}{\textbf{\circled{U}}}}&18.34 & 23.60 & \textbf{0.28} &	\textbf{0.38}&  \textbf{1.55} & \textbf{2.17}\\
		\midrule
		\textbf{Drive\&Act - All Classes} \\
		P3D \cite{qiu2017learning} \scriptsize{\textcolor{table_standard}{\textbf{\circled{S}}}}&17.89 & 21.09 &0.28&	0.34& 1.77 & 2.12 \\
		I3D \cite{carreira2017quo} \scriptsize{\textcolor{table_standard}{\textbf{\circled{S}}}}&11.72 & 15.97 & 0.14 &	0.19&  1.10 & 1.56 \\
		P3D + Temperature Scaling  \cite{guo2017calibration} \scriptsize{\textcolor{table_uncertainty}{\textbf{\circled{U}}}}&5.89 & 6.41 & 0.26&	0.30&  1.35 & 1.63\\
		I3D + Temperature Scaling \cite{guo2017calibration} \scriptsize{\textcolor{table_uncertainty}{\textbf{\circled{U}}}}&6.59 & 8.55 &0.12&	\textbf{0.17}&  0.68 & 0.99\\
		CARING - P3D (ours) \scriptsize{\textcolor{table_uncertainty}{\textbf{\circled{U}}}}&4.58 & \textbf{5.26} & 0.24&	0.30& 1.26 & 1.57 \\
		CARING - I3D (ours) \scriptsize{\textcolor{table_uncertainty}{\textbf{\circled{U}}}}&\textbf{3.03} & 6.02 & \textbf{0.11} & \textbf{0.17}& \textbf{0.58} & \textbf{0.9} \\ 
		\multicolumn{7}{c}{\scriptsize{\textcolor{table_standard}{\textbf{\circled{S}}}} Standard activity recognition models\quad \quad \quad \quad  \scriptsize{\textcolor{table_uncertainty}{\textbf{\circled{U}}}}  Uncertainty-aware models}\\
		
		\bottomrule

	\end{tabular}
		}
		\caption{Reliability of confidence values on the Drive\&Act~\cite{martin2019drive} for standard activity recognition models and their extensions with uncertainty-aware calibration algorithms.
	}
	\vspace{-0.3cm}
	\label{tbl:reliability_results}

\end{table}

For P3D- and I3D models, the input are snippets of $64$ consecutive frames.
If the original video segment is longer, the snippet is randomly chosen during training and at the center of the video at test-time.
If the segment is shorter, the last frame will be repeated until the $64$-frame snippet has been filled.

Following the problem formalization in Section~\ref{sec:definition}, the standard accuracy-driven evaluation protocols~\cite{martin2019drive} have been extended with the expected calibration error (ECE), which depicts the divergence of model confidence score from the true misclassification probability.
Additionally, results under the metric of Negative Log Likelihood (NLL) are presented, as high NLL values are strongly correlated to model miscalibration~\cite{guo2017calibration}.
Overall, we average results over the three spits.

\subsection{Confidence estimates for driver activity recognition}
\label{sec:exp-closed-set}

\begin{table*}[!t]
	\centering

	\scalebox{1.23}{	
\begin{tabular}{lrccccccc}
	\toprule
	\multirow{2}{*}{Activity}       & \multirow{3}{*}{\begin{tabular}[c]{@{}l@{}}Number of\\ Samples\end{tabular}} & \multirow{3}{*}{Recall (\%)} & \multicolumn{3}{c}{I3D \scriptsize{\textcolor{table_standard}{\textbf{\circled{S}}}}}      & \multicolumn{3}{c}{CARING-I3D \scriptsize{\textcolor{table_uncertainty}{\textbf{\circled{U}}}}} \\
	&                                                                              &                         & \begin{tabular}[c]{@{}c@{}}Mean\\ Conf. (\%)\end{tabular} & \begin{tabular}[c]{@{}c@{}}$\Delta$Acc\\(\%)\end{tabular} & \begin{tabular}[c]{@{}c@{}}ECE\\(\%)\end{tabular}   & \begin{tabular}[c]{@{}c@{}}Mean\\ Conf. (\%)\end{tabular}  & \begin{tabular}[c]{@{}c@{}}$\Delta$Acc\\(\%)\end{tabular} & \begin{tabular}[c]{@{}c@{}}ECE\\(\%)\end{tabular}   \\
	\midrule
	\multicolumn{2}{l}{\textbf{Five most common activities}}    &&&&&&&                                                                                                                                                       \\
	sitting\_still                  & 2797                                                                         & 95.1                    & 97.96     & 2.86     & 2.86  & 93.84      & -1.26     & \textbf{1.84}\\
	
	eating                          & 877                                                                          & 86.42                   & 93.26     & 6.84     & 9.33  & 80.99      & -5.43     & \textbf{5.75} \\
	fetching\_an\_object            & 756                                                                          & 76.03                   & 93.77     & 17.74    & 18.28 & 79.42      & 3.4       &\textbf{5.32}  \\
	placing\_an\_object             & 688                                                                          & 66.77                   & 93.03     & 26.25    & 26.25 & 75.9       & 9.13      & \textbf{9.25}  \\
	reading\_magazine               & 661                                                                          & 92.93                   & 98.58     & 5.65     & 6.09  & 93.35      & 0.42      & \textbf{2.87}  \\

	\midrule
	\multicolumn{2}{l}{\textbf{Five most underrepresented activities}}   &&&&&&&                                                                                                                                                   \\
	closing\_door\_inside           & 30                                                                           & 92.31                   & 98.51     & 6.21     & \textbf{8.22}  & 86.00         & -6.31     & 8.30   \\
	closing\_door\_outside          & 22                                                                           & 81.82                   & 93.55     & 11.73    & 20.97 & 86.86      & 5.04      & \textbf{19.81} \\
	opening\_backpack               & 27                                                                           & 0                       & 98.82     & 98.82    & 98.82 & 82.69      & 82.69     & \textbf{82.69} \\
	putting\_laptop\_into\_backpack & 26                                                                           & 16.67                   & 92.67     & 76.00       & 76.00    & 76.46      & 59.8      & \textbf{59.80}  \\
	taking\_laptop\_from\_backpack  & 19                                                                           & 0.00                       & 85.25     & 85.25    & 85.25 & 70.08      & 70.08     & \textbf{70.08}\\
	\multicolumn{9}{c}{\scriptsize{\textcolor{table_standard}{\textbf{\circled{S}}}} Standard activity recognition models\quad \quad \quad \quad  \scriptsize{\textcolor{table_uncertainty}{\textbf{\circled{U}}}}  Uncertainty-aware models}\\
	\bottomrule
\end{tabular}
	}
	\caption{
	Analysis of the resulting confidence estimates of the initial I3D model and its CARING version for individual common and rare Drive\&Act activities. \emph{Recall} denotes the recognition accuracy of the current class, while \emph{Mean Conf.} denotes the average confidence estimate produced by the model. Supplemental to the Expected Calibration Error (\emph{ECE}), we report the difference between the mean confidence value and model accuracy (denoted \emph{$\Delta$Acc}). While in a perfectly calibrated model $\Delta$Acc is 0, ECE is a better evaluation metric, as \eg~if a lot of samples have too high and too low confidence values, their average might lead to a misconception of good calibration. While there is room for improvement for underrepresented and poorly recognized activity classes, the CARING model  leads to better uncertainty estimates. 
}
	
	\label{tbl:class-wise}
	
\end{table*}

In Table~\ref{tbl:reliability_results}, we compare the described CNN-based activity recognition approaches among themselves and with their uncertainty-aware versions in terms of ECE, Brier Score, and NLL for \emph{overepresented}, \emph{rare}, and \emph{all} classes in Drive\&Act.
First, the conjecture has been confirmed that native activity recognition architectures yield unreliable confidence predictions. 
Evidently, confidence scores produced by I3D score have a misalignment of $15.97\%$ for Drive\&Act.
Similar problems have been observed in P3D: $21.2\%$ ECE on Drive\&Act, an exceedingly high error for safety-critical driving applications.


Model reliability has been clearly improved by learning to attain proper probability estimates, since all of the uncertainty-aware versions exceed the original Softmax values.
To our surprise, while I3D has better initial uncertainty estimates than P3D, \ie, $15.97\%$ for I3D in contrast with ECE of $21.09\%$ for P3D, it seems that P3D has a stronger response to both, temperature scaling and CARING approaches than I3D, resulting ECE of $5.26\%$ for CARING-P3D and $6.02\%$ for CARING-I3D.
However, this difference is less than $1\%$, and thereby one would advocate to use I3D due to its higher accuracy in most cases~\cite{martin2019drive,carreira2017quo,qiu2017learning}.
While the expected calibration error is deemed to be essentially important for real applications, one must realize that this metric is complementary to model accuracy and both measures should be taken into consideration when selecting the suitable model for deployment.
We further mark, that both temperature scaling and the CARING method \emph{do not influence the model accuracy}, as illustrated in Sections \ref{sec:temperature} and \ref{sec:caring}. 
Regarding Pseudo 3D ResNet, we have achieved an overall accuracy of $54.86\%$ on the validation split and $46.62\%$ on the test split of Drive\&Act, which does not change after our uncertainty-inspired adaptations.
In line with~\cite{martin2019drive}, I3D reaches a higher accuracy of $68.71\%$ on the validation set and $63.09\%$ on the test set.\footnote{The slight deviation from the accuracy reported in the original work\cite{martin2019drive} (between $0.18\%$ and $1.3\%$) is due to random factors in the training process.}

It arrives as expected, that the model confidence reliability is correlated to the amount of training data, as seen in the distinguished areas for \emph{common}, \emph{rare}, and \emph{all} classes of Drive\&Act (Table~\ref{tbl:reliability_results}).
A case in point lies in the \emph{common classes} setting, which encounters the lowest expected calibration error for both original and uncertainty-aware architectures, \ie, $13.38\%$ for I3D and $5.26\%$ for CARING-I3D.
Using intermediate input representation through our CARING calibration network results in the best probability estimates on both datasets and in all evaluation settings.
More precisely, with the proposed CARING strategy, our model clearly stands out in front of the raw neural network and the temperature scaling-based one, surpassing them by $9.95\%$ and $2.53\%$ on Drive\&Act.
The differences are smaller when considering the Brier Score metric, presumably because it is especially insensitive to rare categories~\cite{lakshminarayanan2017simple}, which, however, might be of high importance in driver observation, as uncommon events are often linked to distraction.
Still, also in terms of the Brier score, the advantages of the CARING model are evident, while the largest performance gain is achieved for underrepresented classes.
Overall, our experiments highlight the benefits of learning to attain probability scores \emph{depending on the input}.

\begin{figure}[]
	\centering
	\includegraphics[width=0.95\linewidth ]{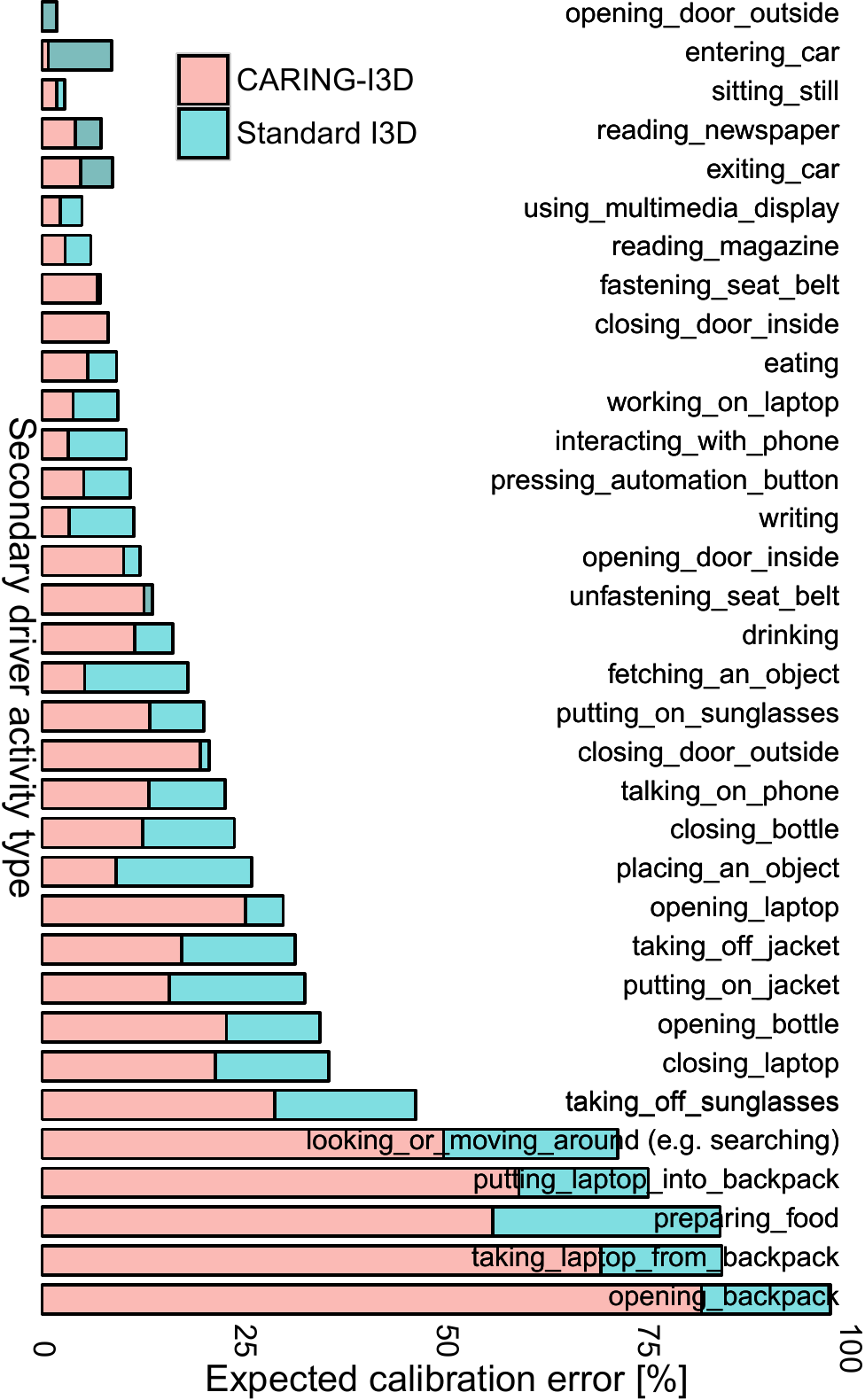}
	\caption{\revised{Exepcted Calibration Error (ECE) for all 34 secondary activity categories of the Drive\&Act dataset (validation split).}}
	\label{fig:results_all_classes}
		\vspace{-0.3cm}
\end{figure}

Moreover, we assess model performance for the individual classes by examining the top five common and the five least frequent activities in Drive\&Act, whose results are displayed in Table~\ref{tbl:class-wise} in a separate manner.
In addition to ECE, we include the accuracy for samples belonging to the individual classes, the average confidence value achieved with the corresponding model, and the difference between them (denoted as \emph{$\Delta$Acc}).
While a such global confidence-accuracy disagreement is interesting to consider (and is $0$ for a perfectly calibrated model), it should be interpreted with care, as it may bring about an imprecise illusion of good confidence calibration, as \eg, many samples with too high and too low confidence values may cancel each other out due to the averaging effects.

Reliability of the confidence scores has been significantly improved via the CARING method and is linked to the amount of training samples and the accuracy.
Models have clear issues when learning with few examples, and have a limited performance, \eg, $76.00\%$ for I3D and $59.80\%$ for CARING-I3D ECE in dealing with \emph{putting laptop into backpack}.
In contrast, for both, over- and underrepresented classes, the ECE of easy-to-identify activities (\ie, the ones with high accuracy) is much lower.
Before calibrated, the average confidence value is always higher than the accuracy (positive $\Delta Acc$), revealing that the models are too optimistic in their predictions.
Interestingly, after the CARING conversion is executed, the average model confidence becomes lower than the accuracy for some classes, such as \emph{eating}.
In this sense, CARING models are more conservative in their assessment of prediction certainty.
\looseness=-1

\subsection{\revised{Results for different driver behaviours}}
\label{sec:exp-different}

\revised{
We now look at different driver behaviour types individually, with the exact ECE for the original classifier and our CARING model provided in Figure \ref{fig:results_all_classes}. We observe in Figure \ref{fig:results_all_classes}, that  nearly all secondary behaviours exhibit better quality of the confidence estimates with the proposed CARING model, but certain behaviours still have a large room for improvement.}

\revised{
As previously validated in Table~\ref{tbl:class-wise}, the realism of confidences is especially good for more discriminative behaviors (which also obtain higher accuracies) and classes with  sufficient training examples. 
For instance, the activity \textit{eating}  has around 800 examples in Drive\&Act~\cite{martin2019drive} and  a much lower ECE compared to a similar category \textit{preparing food}, of which there are only $\sim 50$ examples. 
Especially a mixture of low distinctiveness (\textit{e.g.}, \textit{opening backpack} vs. \textit{taking laptop from backpack}) and low number of training examples are fatal for a category: \textit{taking laptop backpack} is off by $70\%$ in terms of ECE (with CARING), which goes up to $85\%$ without additional confidence calibration. 
In other words, the model did not correctly predict the class in nearly all cases of this behaviour, while the estimated confidence was very high. 
}

\subsection{Observation results from alternate views and modalities}
\label{sec:exp-views}

All activities in Drive\&Act are captured by a set of six cameras, five lightweight  near-infrared (NIR) cameras and one Kinect camera.
The NIR cameras covers several different views from the interior of the vehicle, namely: \emph{front top}, \emph{right top}, \emph{back}, \emph{face view}, and \emph{left top}.
In addition, the Kinect camera provides three different  modalities: near-infrared, color,  and depth.
All cameras are synchronized at the frame level, therefore every labeled activity is captured by every camera and corresponds to the same time interval.
The principal modality, used in the other evaluation tasks, corresponds to the \emph{front top} view, as done in~\cite{martin2019drive}.

In Table~\ref{tbl:multi-modal}, we show the performance of CARING in all available views and modalities.
One of the key differences between them is the difficulty of the task, \eg, it is easier to recognize an activity in views where the entire body is visible than in views where only the face is visible.
We can observe how the \emph{face view}, with its limited aperture, obtains a mean accuracy amongst all classes of only $49.57\%$, whereas the \emph{front top} obtains an accuracy of $68.71\%$.
Therefore, we can compare how CARING behaves when recognizing the same activities from different levels of difficulty.
The ECE of the raw CNN hovers between $10\%$ and $20\%$  on most of the modalities, albeit it shows significant variance and rises up to $80.45\%$ on the \emph{Kinect IR} modality when only common cases are considered.
Conversely, the ECE of the CARING models not only is lower, but it shows a very stable behavior with little variance between different views and modalities.
\revised{Compared to the raw 3D CNN and temperature scaling approach on the main sensor modality of Drive\&Act~\cite{martin2019drive}, \textit{i.e.}, Front top NIR, CARING improves the validation and test sets ECE by $13.14\%$ and $19.72\%$ for rare classes, $7.65\%$ and $7.16\%$ for common classes and $8.68\%$ and $9.96\%$ for all classes, indicating clear advantages of the proposed CARING model for uncertainty estimation, which is consistent for different amounts of training data (although the overall confidence quality declines with less training examples for all considered approaches). 
Similar results are observed for other Drive\&Act modalities and sensor placements.
For example, for the  Kinect Depth modality, CARING outperforms the other two methods on validation and test sets by $12.12\%$ and $16.28\%$ for rare classes, $7.07\%$ and $7.10\%$ for common classes and $5.91\%$ and $9.02\%$ for all classes, providing empirical evidence for cross-modal generalization abilities of CARING.}
Even so, we observe that the ECE of the most challenging view is larger than the ECE of less challenging views, which is consistent with a confidence model where the network is more self-confident on easier tasks.


\begin{table*}[]
\resizebox{\linewidth}{!}{%
\begin{tabular}{lcccccccc|cccccccc|cccccccc} 
\toprule
Model & \multicolumn{8}{c}{Rare classes} & \multicolumn{8}{c}{Common classes} & \multicolumn{8}{c}{All classes} \\
 & \multicolumn{2}{c}{ECE (\%)} & \multicolumn{2}{c}{Brier} & \multicolumn{2}{c}{NLL} & \multicolumn{2}{c}{Acc (\%)} & \multicolumn{2}{c}{ECE (\%)} & \multicolumn{2}{c}{Brier} & \multicolumn{2}{c}{NLL} & \multicolumn{2}{c}{Acc (\%)} & \multicolumn{2}{c}{ECE (\%)} & \multicolumn{2}{c}{Brier} & \multicolumn{2}{c}{NLL} & \multicolumn{2}{c}{Acc (\%)} \\
 & val & test & val & test & val & test & val & \multicolumn{1}{c}{test} & val & test & val & test & val & test & val & \multicolumn{1}{c}{test} & val & test & val & test & val & test & val & test \\ 
\midrule
\textbf{Front top NIR} &  &  &  &  &  &  &  &  &  &  &  &  &  &  &  &  &  &  &  &  &  &  &  &  \\
Raw 3D CNN & 31.48 & 43.32 & 0.34 & 0.47 & 3.41 & 4.54 & \multirow{3}{*}{57.02}   & \multirow{3}{*}{48.68}  & 10.22 & 13.38 & 0.12 & 0.16 & 0.90 & 1.27 & \multirow{3}{*}{80.41
} & \multirow{3}{*}{77.5
} & 11.72 & 15.98 & 0.14 & 0.19 & 0.90 & 1.56 &\multirow{3}{*}{68.71
} & \multirow{3}{*}{63.09
} \\
Temp. Scaling & 24.97 & 32.38 & 0.30 & 0.41 & 1.96 & 2.62 &  &  & 5.31 & 6.99 & 0.11 & 0.15 & 0.57 & 0.83 &  &  & 6.59 & 8.55 & 0.12 & 0.17 & 0.68 & 0.99 &  &  \\
CARING & 18.34 & 23.60 & 0.28 & 0.38 & 1.55 & 2.17 &  &  & 2.57 & 6.22 & 0.10 & 0.15 & 0.50 & 0.78 &  &  & 3.04 & 6.02 & 0.11 & 0.17 & 0.58 & 0.90 &  &  \\ 
\midrule
\textbf{Right top NIR} &  &  &  &  &  &  &  &  &  &  &  &  &  &  &  &  &  &  &  &  &  &  &  &  \\
Raw 3D CNN & 35.89 & 42.86 & 0.40 & 0.48 & 3.79 & 4.78 &  \multirow{3}{*}{55.57}& \multirow{3}{*}{49.78
} & 14.73 & 15.83 & 0.18 & 0.19 & 1.28 & 1.42 &  \multirow{3}{*}{72.31}& \multirow{3}{*}{70.06}  & 16.26 & 18.23 & 0.19 & 0.22 & 1.48 & 1.73 &  \multirow{3}{*}{63.94}& \multirow{3}{*}{59.92} \\
Temp. Scaling & 25.69 & 29.52 & 0.35 & 0.41 & 2.20 & 2.73 &  &  & 6.59 & 6.57 & 0.15 & 0.17 & 0.82 & 0.93 &  &  & 7.71 & 7.97 & 0.17 & 0.20 & 0.93 & 1.10 &  &  \\
CARING & 20.49 & 24.03 & 0.33 & 0.40 & 1.92 & 2.39 &  &  & 3.39 & 5.30 & 0.15 & 0.17 & 0.76 & 0.90 &  &  & 3.67 & 5.67 & 0.16 & 0.19 & 0.85 & 1.03 &  &  \\ 
\midrule
\textbf{Back NIR} &  &  &  &  &  &  &  &  &  &  &  &  &  &  &  &  &  &  &  &  &  &  &  &  \\
Raw 3D CNN & 40.17 & 49.47 & 0.46 & 0.54 & 4.42 & 5.23 &  \multirow{3}{*}{44.9}& \multirow{3}{*}{39.95}  & 16.07 & 19.55 & 0.22 & 0.23 & 1.58 & 1.67 & \multirow{3}{*}{65.18}& \multirow{3}{*}{66.89}  & 17.85 & 22.14 & 0.24 & 0.26 & 4.42 & 2.00 &  \multirow{3}{*}{55.04}& \multirow{3}{*}{53.42}  \\
Temp. Scaling & 25.45 & 33.69 & 0.39 & 0.46 & 2.70 & 3.00 &  &  & 5.19 & 9.00 & 0.20 & 0.20 & 1.14 & 1.06 &  &  & 6.03 & 11.00 & 0.21 & 0.22 & 1.26 & 1.24 &  &  \\
CARING & 19.95 & 26.58 & 0.37 & 0.43 & 2.41 & 2.67 &  &  & 3.48 & 5.25 & 0.19 & 0.19 & 1.07 & 1.01 &  &  & 3.17 & 6.59 & 0.20 & 0.21 & 1.17 & 1.16 &  &  \\ 
\midrule
\textbf{Face View NIR} &  &  &  &  &  &  &  &  &  &  &  &  &  &  &  &  &  &  &  &  &  &  &  &  \\
Raw 3D CNN & 39.22 & 45.56 & 0.47 & 0.54 & 4.60 & 4.97 &  \multirow{3}{*}{42.09}& \multirow{3}{*}{34.92}  & 19.89 & 21.40 & 0.29 & 0.31 & 2.02 & 2.02 &  \multirow{3}{*}{57.05}& \multirow{3}{*}{49.73}  & 21.34 & 23.50 & 0.30 & 0.33 & 2.22 & 2.28 &  \multirow{3}{*}{49.57}& \multirow{3}{*}{42.32}  \\
Temp. Scaling & 24.76 & 26.14 & 0.40 & 0.45 & 2.59 & 3.03 &  &  & 6.30 & 6.68 & 0.26 & 0.28 & 1.39 & 1.53 &  &  & 6.78 & 6.81 & 0.27 & 0.30 & 1.48 & 1.67 &  &  \\
CARING & 18.93 & 23.20 & 0.38 & 0.44 & 2.37 & 2.94 &  &  & 4.22 & 10.32 & 0.25 & 0.28 & 1.27 & 1.53 &  &  & 4.63 & 9.87 & 0.26 & 0.30 & 1.36 & 1.66 &  &  \\ 
\midrule
\textbf{Left top NIR} &  &  &  &  &  &  &  &  &  &  &  &  &  &  &  &  &  &  &  &  &  &  &  &  \\
Raw 3D CNN & 31.17 & 44.11 & 0.34 & 0.48 & 2.96 & 4.23 &  \multirow{3}{*}{52.64}& \multirow{3}{*}{47.93}   & 9.43 & 12.43 & 0.11 & 0.16 & 0.92 & 1.18 &  \multirow{3}{*}{80.74}& \multirow{3}{*}{75.80} & 11.10 & 15.16 & 0.13 & 0.19 & 1.08 & 1.46 &  \multirow{3}{*}{66.69}& \multirow{3}{*}{61.87}  \\
Temp. Scaling & 21.46 & 31.12 & 0.30 & 0.41 & 1.79 & 2.58 &  &  & 4.96 & 6.23 & 0.10 & 0.15 & 0.59 & 0.83 &  &  & 5.88 & 7.69 & 0.12 & 0.17 & 0.69 & 0.99 &  &  \\
CARING & 18.14 & 25.46 & 0.28 & 0.40 & 1.53 & 2.34 &  &  & 2.45 & 6.58 & 0.10 & 0.15 & 0.49 & 0.82 &  &  & 2.95 & 5.77 & 0.11 & 0.17 & 0.57 & 0.97 &  &  \\ 
\midrule
\textbf{Kinect IR} &  &  &  &  &  &  &  &  &  &  &  &  &  &  &  &  &  &  &  &  &  &  &  &  \\
Raw 3D CNN & 29.65 & 36.12 & 0.29 & 0.45 & 3.03 & 3.69 &  \multirow{3}{*}{63.83}& \multirow{3}{*}{53.16}  & 80.45 & 13.69 & 0.12 & 0.15 & 0.84 & 1.253 &  \multirow{3}{*}{82.10}& \multirow{3}{*}{76.94}  & 9.51 & 15.67 & 0.13 & 0.18 & 1.01 & 1.48 &  \multirow{3}{*}{72.96}& \multirow{3}{*}{65.05}  \\
Temp. Scaling & 20.81 & 23.88 & 0.26 & 0.39 & 2.00 & 2.38 &  &  & 49.95 & 6.17 & 0.11 & 0.14 & 0.64 & 0.89 &  &  & 5.11 & 7.49 & 0.12 & 0.16 & 0.74 & 1.03 &  &  \\
CARING & 15.66 & 19.90 & 0.25 & 0.37 & 1.78 & 2.20 & & & 2.81 & 5.19 & 0.10 & 0.14 & 0.53 & 0.85 &   &  & 2.69 & 6.16 & 0.11 & 0.16 & 0.63 & 0.97 &  &  \\ 
\midrule
\textbf{Kinect Depth} &  &  &  &  &  &  &  &  &  &  &  &  &  &  &  &  &  &  &  &  &  &  &  &  \\
Raw 3D CNN & 32.43 & 41.69 & 0.34 & 0.47 & 3.23 & 4.44 &  \multirow{3}{*}{61.34}& \multirow{3}{*}{48.80}   & 11.72 & 16.83 & 0.15 & 0.21 & 1.01 & 1.51 &  \multirow{3}{*}{77.47}& \multirow{3}{*}{70.86}  & 13.13 & 18.94 & 0.16 & 0.23 & 1.17 & 1.78 &  \multirow{3}{*}{69.40}& \multirow{3}{*}{59.83}  \\
Temp. Scaling & 23.88 & 29.36 & 0.30 & 0.41 & 1.94 & 2.66 &  &  & 4.93 & 6.81 & 0.14 & 0.19 & 0.70 & 1.03 &  &  & 5.82 & 8.32 & 0.15 & 0.21 & 0.80 & 1.18 &  &  \\
CARING & 18.84 & 25.05 & 0.28 & 0.39 & 1.67 & 2.35 &  &  & 3.08 & 5.00 & 0.13 & 0.18 & 0.65 & 0.97 &  &  & 3.44 & 5.50 & 0.14 & 0.20 & 0.72 & 1.10 &  &  \\ 
\midrule
\textbf{Kinect Color} &  &  &  &  &  &  &  &  &  &  &  &  &  &  &  &  &  &  &  &  &  &  &  &  \\
Raw 3D CNN & 28.00 & 40.38 & 0.35 & 0.42 & 3.14 & 4.66 &  \multirow{3}{*}{56.42}& \multirow{3}{*}{53.40}  & 9.80 & 13.21 & 0.12 & 0.18 & 0.98 & 1.20 &  \multirow{3}{*}{80.27}& \multirow{3}{*}{72.00}   & 8.99 & 15.56 & 0.14 & 0.20 & 1.14 & 1.51 &  \multirow{3}{*}{68.34}& \multirow{3}{*}{62.70}  \\
Temp. Scaling & 21.47 & 31.06 & 0.31 & 0.37 & 1.80 & 2.66 &  &  & 4.93 & 7.64 & 0.12 & 0.16 & 0.60 & 0.78 &  &  & 5.86 & 9.43 & 0.13 & 0.18 & 0.69 & 0.95 &  &  \\
CARING & 15.88 & 24.10 & 0.30 & 0.35 & 1.80 & 2.21 &  &  & 2.73 & 6.10 & 0.11 & 0.16 & 0.51 & 0.74 &  &  & 3.08 & 6.54 & 0.12 & 0.18 & 0.59 & 0.87 &  &  \\
\bottomrule
\end{tabular}
}
\caption{Impact of the confidence adjustment through temperature scaling and our CARING model for different modalities and views inside the vehicle cabin. \revised{\textit{Raw 3D CNN} indicates I3D without temperature scaling, \textit{Temp. Scaling} indicates I3D equipped with the method of ~\cite{guo2017calibration} and \textit{CARING} is our input-guided framework.}
Additionally to the confidence quality metrics (ECE, Brier Score and NLL, where ECE is considered the most reliable measure), we report the classification accuracies of the indivudual modalities for the I3D network on the Drive\&Act dataset. \revised{We evaluate our model not only on all classes but also on rare classes (underreprensented) and common classes (overreprensented).}
Keep in mind, that as both models do not impact the order of the logits, the \textit{accuracy of the model does not change}. In the vast majority of modalities, our confidence improvement through CARING leads to significantly more realistic confidence estimates, \revised{indicating its effectiveness across different sensor types and camera placements inside the vehicle cabin}. }
\label{tbl:multi-modal}
\end{table*}

\subsection{Distribution analysis}
\label{sec:calib_diag}

\begin{figure}
    \centering
    \includegraphics[width=\linewidth]{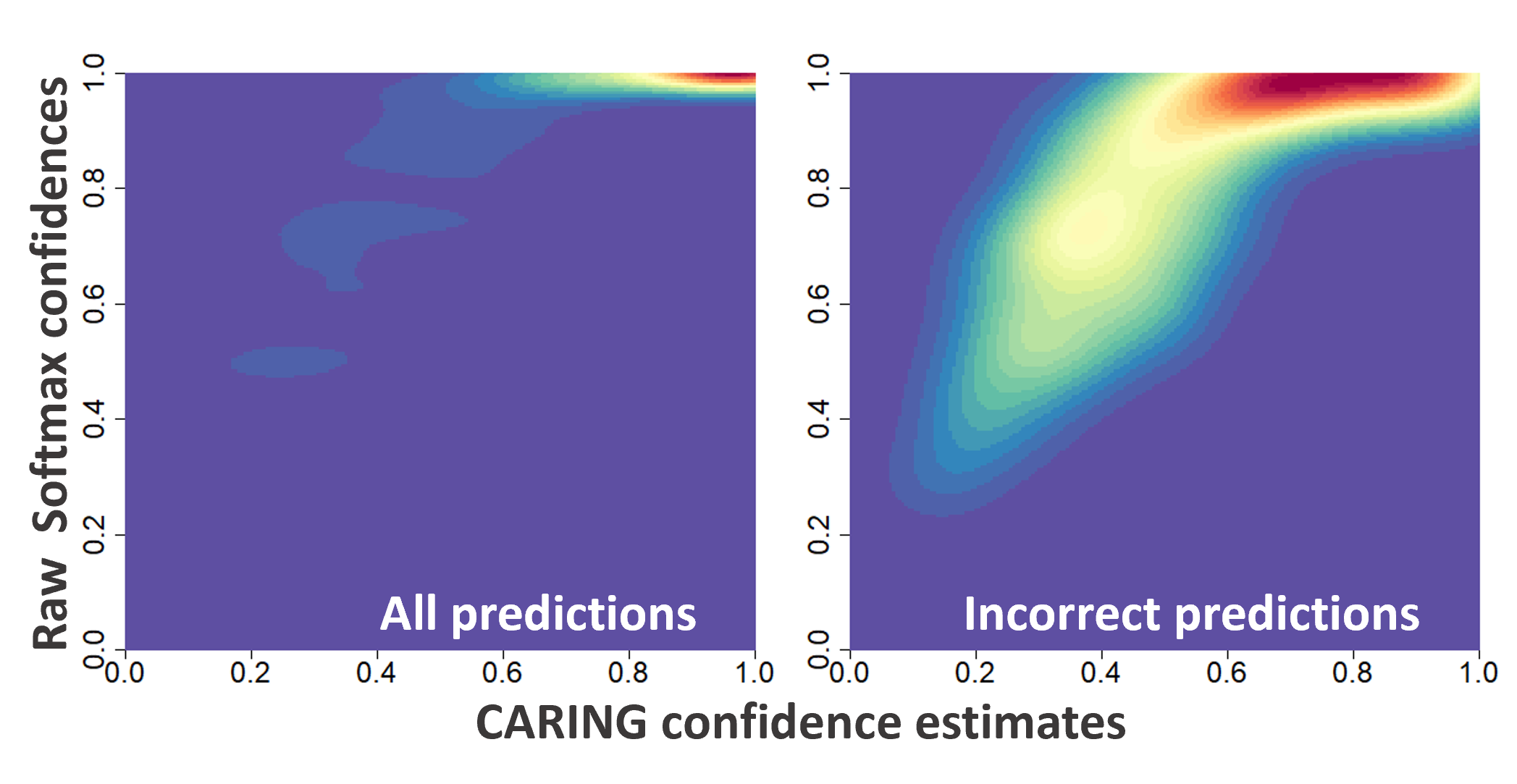}
    \caption{Distribution of predicted confidences as a 2D histogram  for correct and incorrect predictions.
    The Y axis represents standard Softmax probability estimates, while the X axis are confidences improved through the CARING network. Red denotes common cases (high frequency), while blue illustrates unlikely cases.}
    \label{fig:density_confidence}
\end{figure}
\begin{figure*}[]
	\centering
	\begin{subfigure}{.16\textwidth}
		\centering
		\includegraphics[trim={0.1cm 0 1.4cm 0},clip,width=\linewidth]{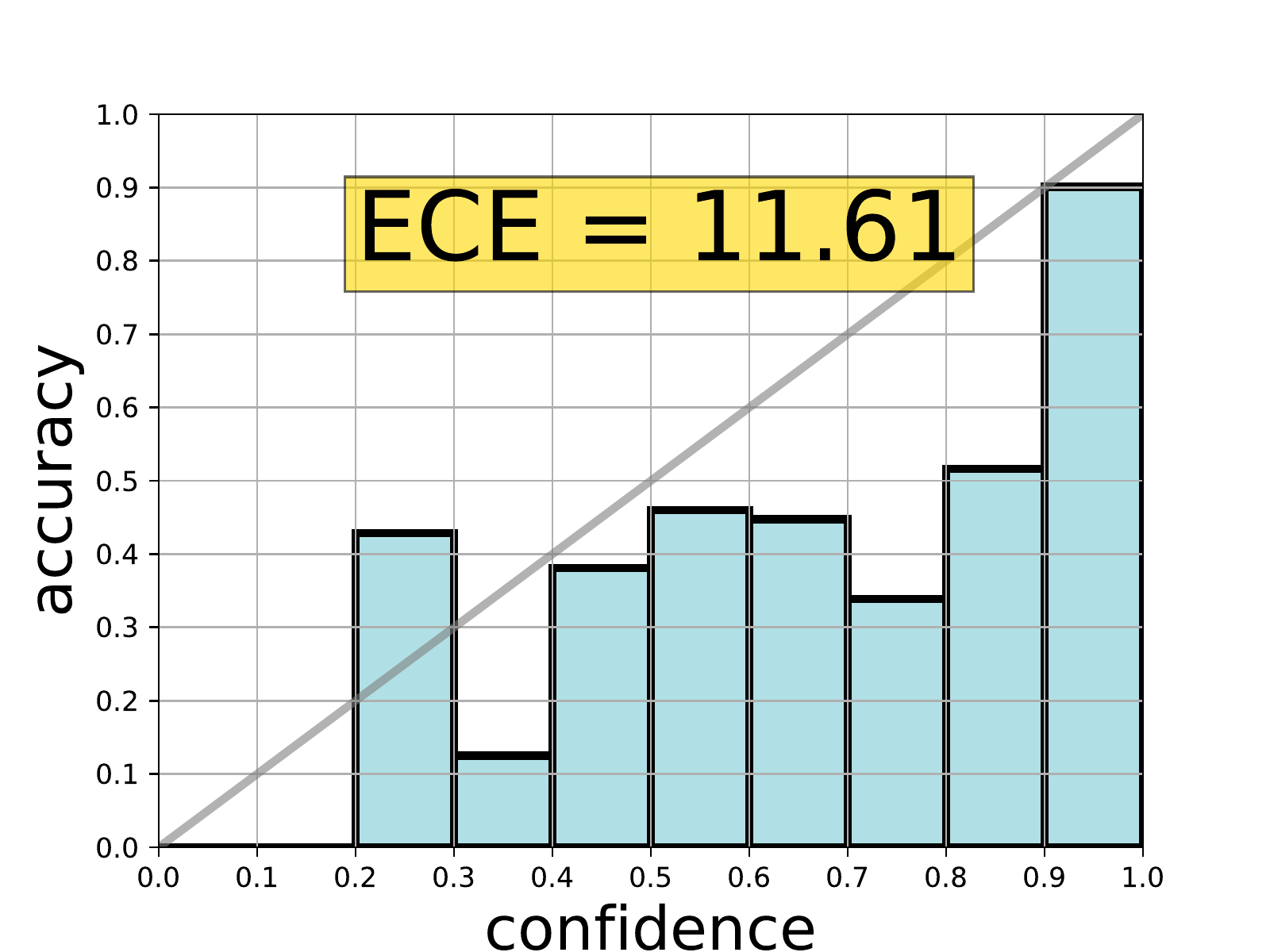}
		\caption{I3D (original), \\ \null \quad all  action classes}
		\label{fig:i3d_all}
	\end{subfigure}%
	\begin{subfigure}{.16\textwidth}
		\centering
		\includegraphics[trim={0.1cm 0 1.4cm 0},clip,width=\linewidth]{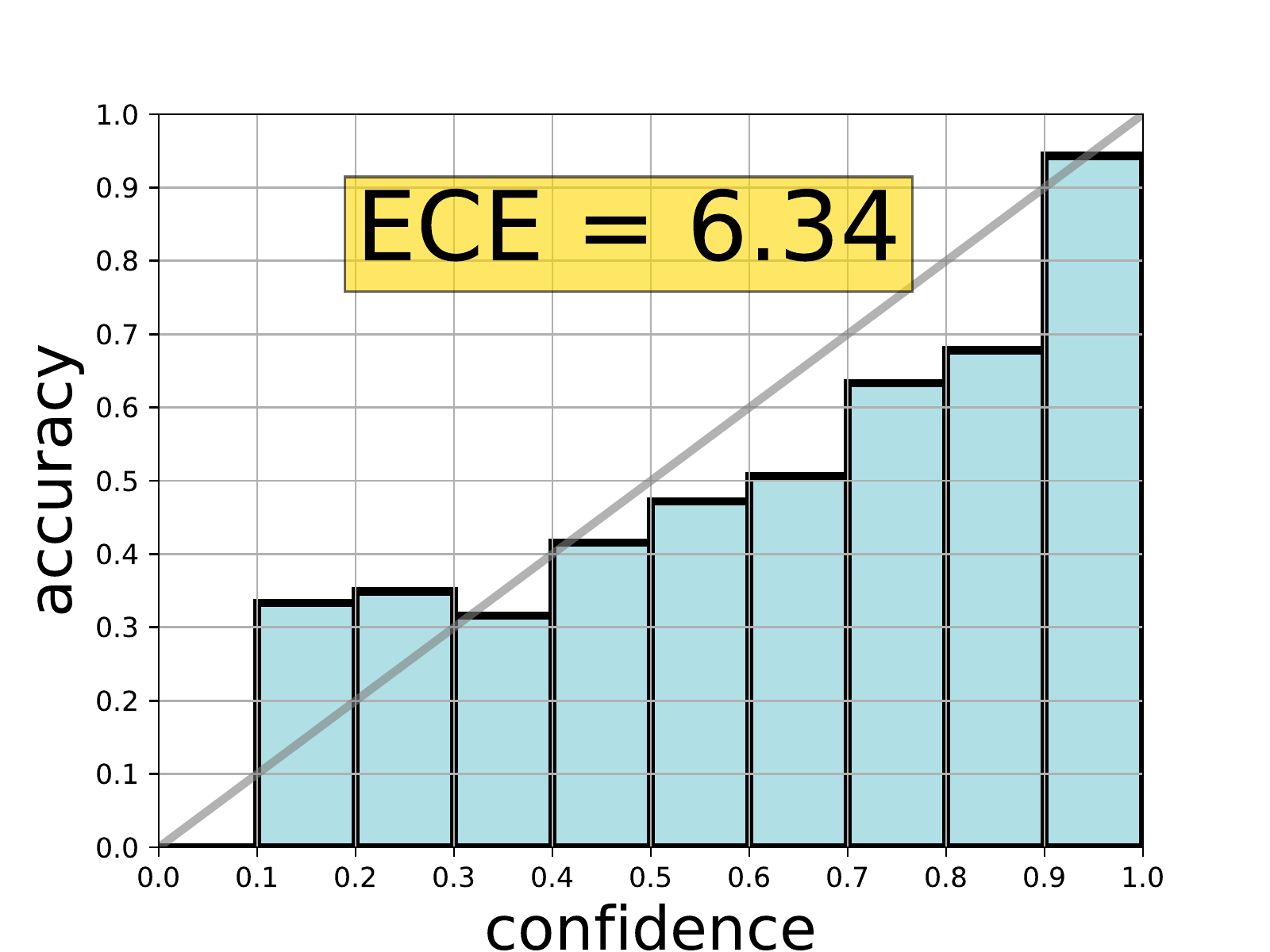}
		\caption{I3D + temp. scaling, all action classes}
		\label{fig:i3d_temp_all}
	\end{subfigure}%
	\begin{subfigure}{.16\textwidth}
		\centering
		\includegraphics[trim={0.1cm 0 1.4cm 0},clip,width=\linewidth]{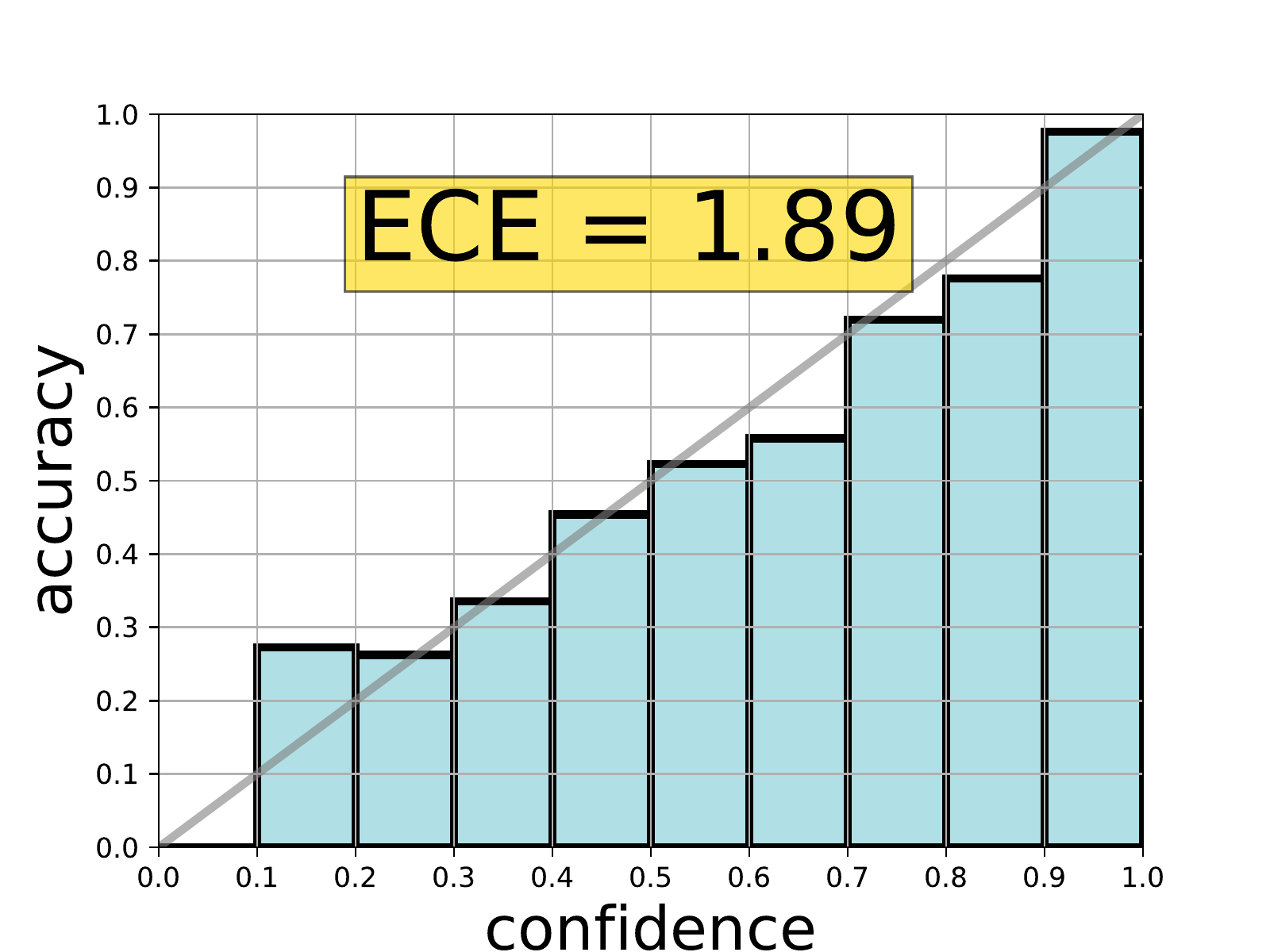}
		\caption{CARING-I3D, \\ \null \quad all action classes}
		\label{fig:i3d_caring_all}
	\end{subfigure}%
	\begin{subfigure}{.16\textwidth}
		\centering
		\includegraphics[trim={0.1cm 0 1.4cm 0},clip,width=\linewidth]{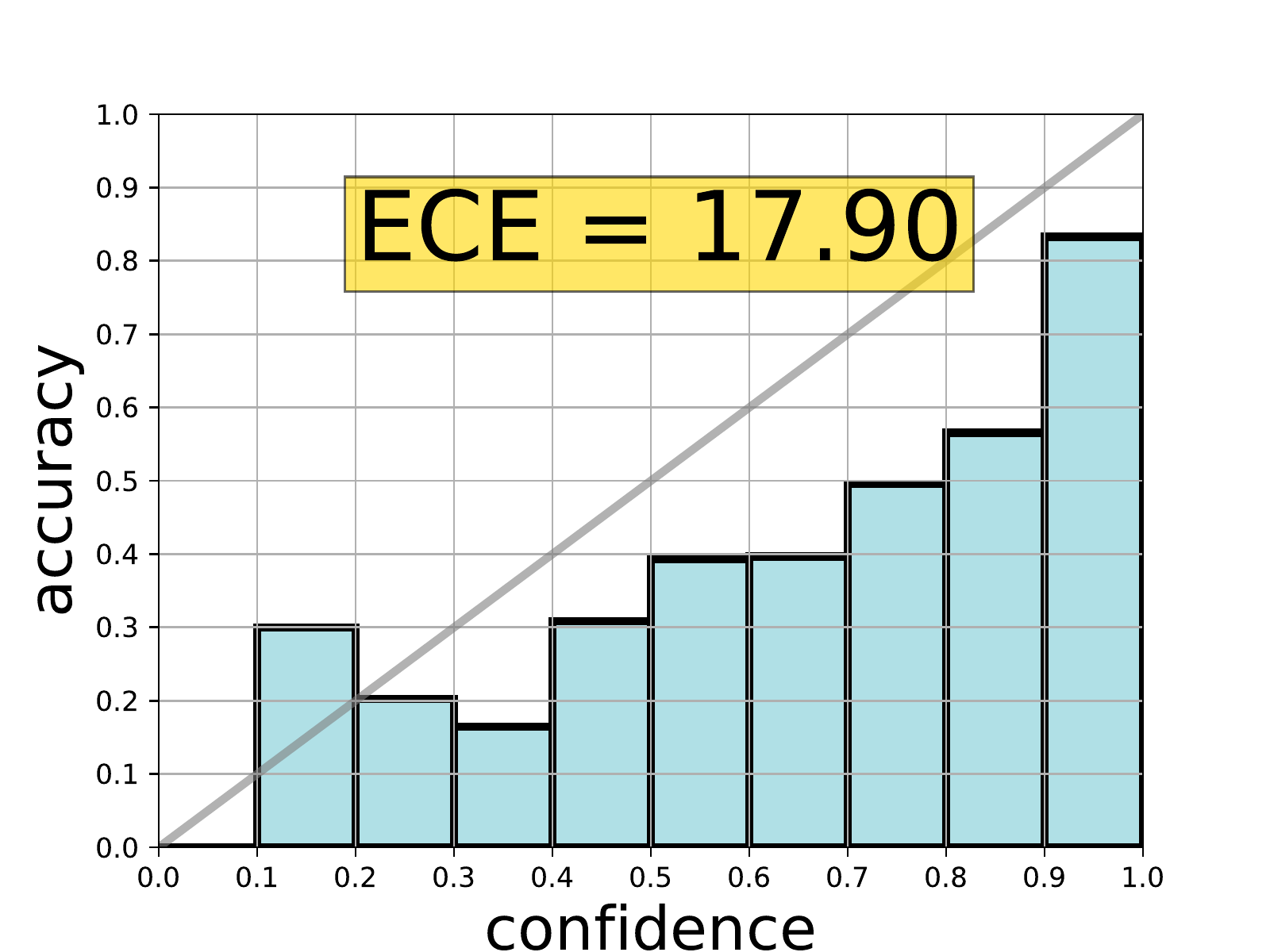}
		\caption{P3D (original), \\ \null \quad all action classes}
		\label{fig:p3d_all}
	\end{subfigure}%
	\begin{subfigure}{.16\textwidth}
		\centering
		\includegraphics[trim={0.1cm 0 1.4cm 0},clip,width=\linewidth]{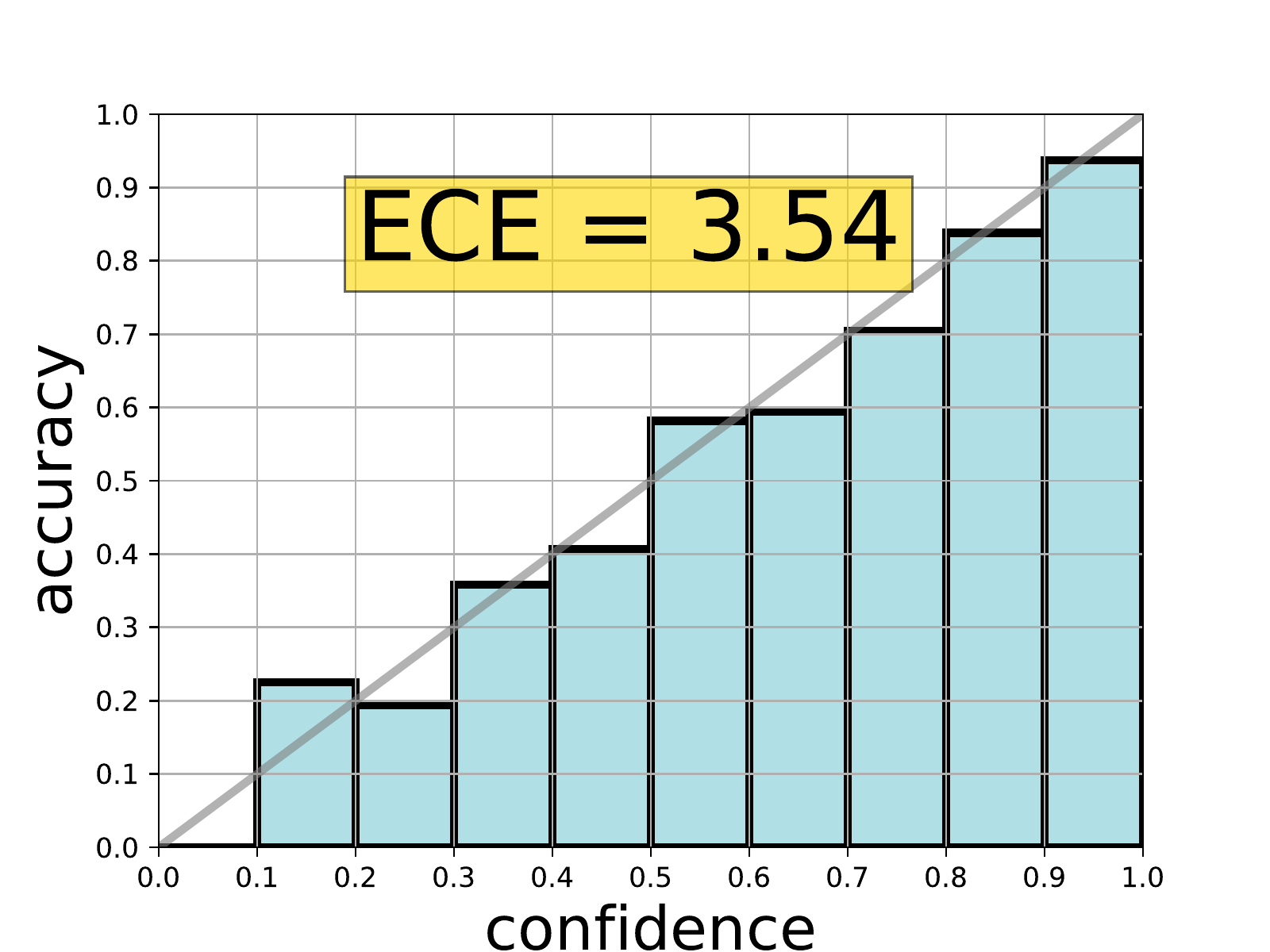}
		\caption{P3D + temp. scaling, all action classes}
		\label{fig:p3d_temp_all}
	\end{subfigure}%
	\begin{subfigure}{.16\textwidth}
		\centering
		\includegraphics[trim={0.1cm 0 1.4cm 0},clip,width=\linewidth]{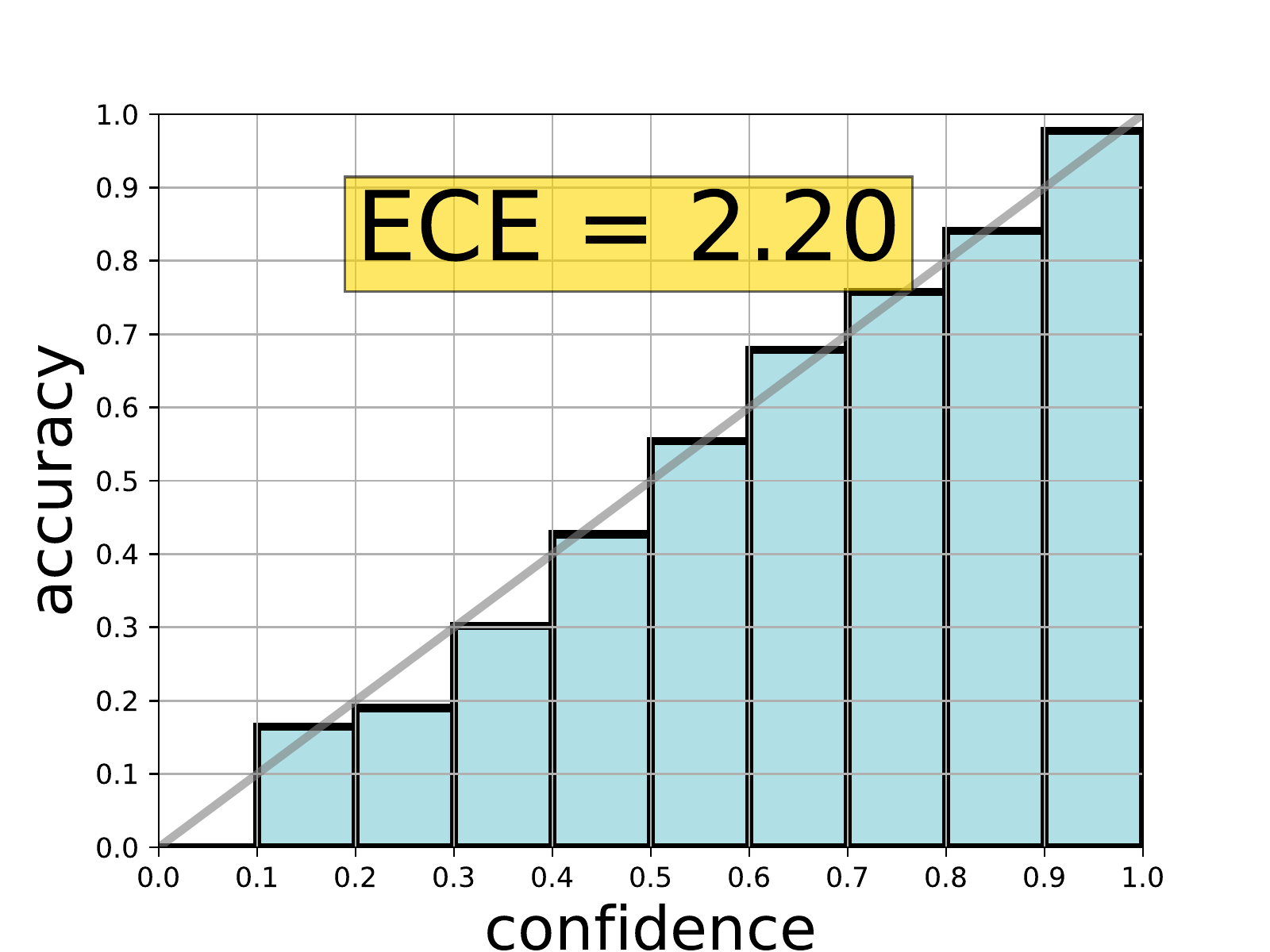}
		\caption{CARING-P3D, \\ \null \quad all action classes}
		\label{fig:p3d_caring_all}
	\end{subfigure}%
	
	\begin{subfigure}{.16\textwidth}
		\centering
		\includegraphics[trim={0.1cm 0 1.4cm 0},clip,width=\linewidth]{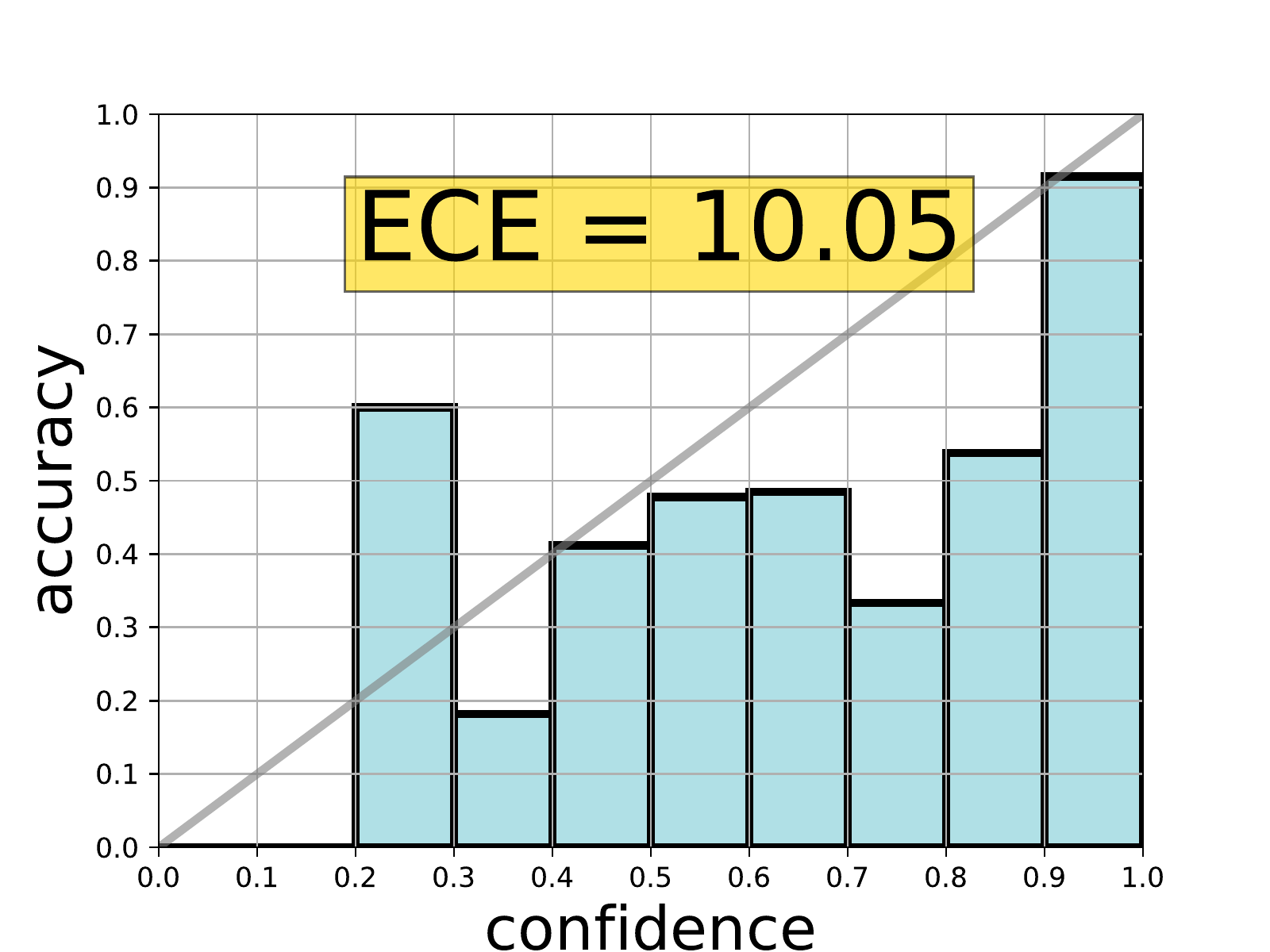}
		\caption{I3D (original), \\ \null \quad common classes}
		\label{fig:i3d_common}
	\end{subfigure}%
	\begin{subfigure}{.16\textwidth}
		\centering
		\includegraphics[trim={0.1cm 0 1.4cm 0},clip,width=\linewidth]{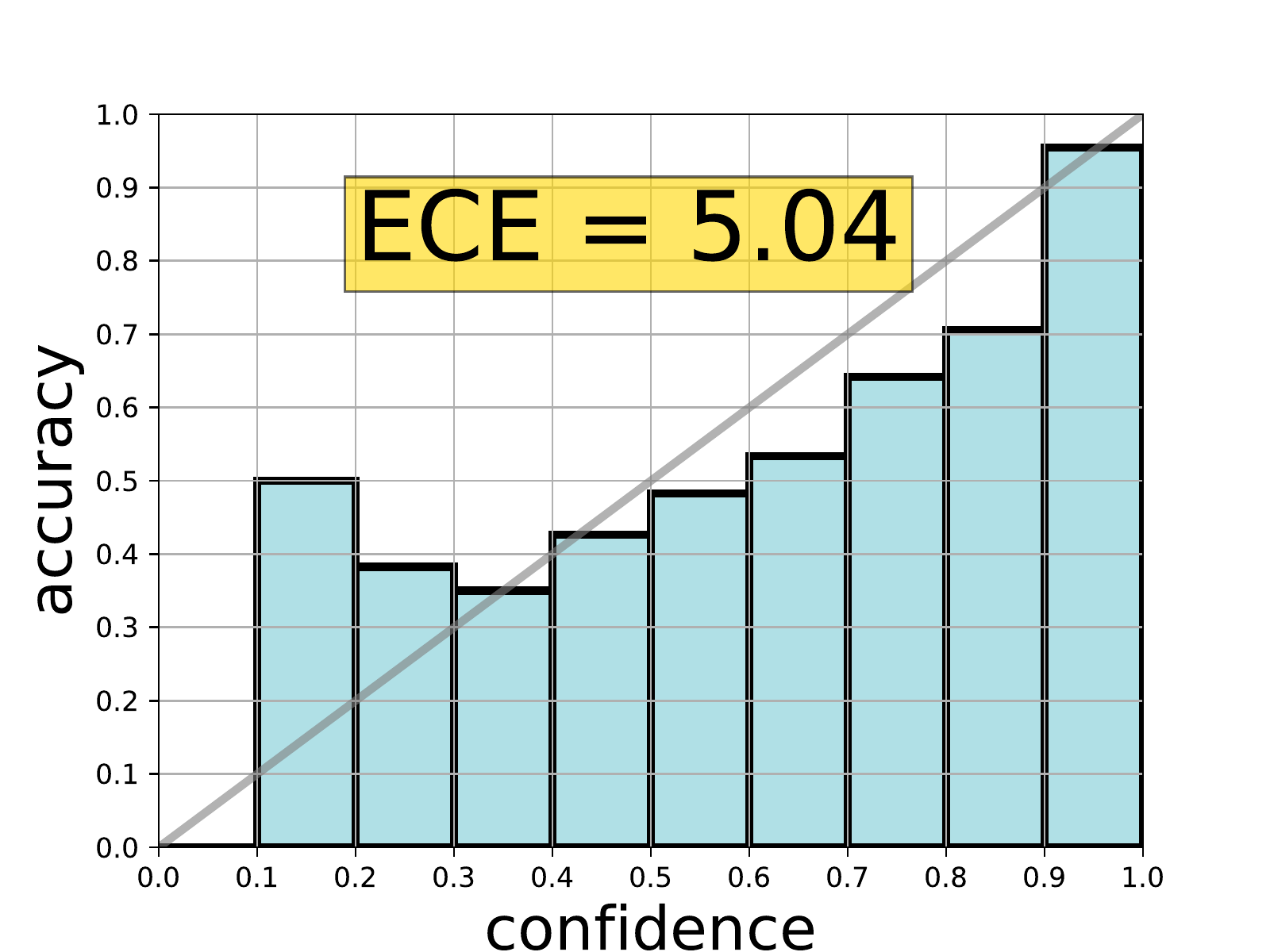}
		\caption{I3D + temp. scaling, common classes}
		\label{fig:i3d_temp_common}
	\end{subfigure}%
	\begin{subfigure}{.16\textwidth}
		\centering
		\includegraphics[trim={0.1cm 0 1.4cm 0},clip,width=\linewidth]{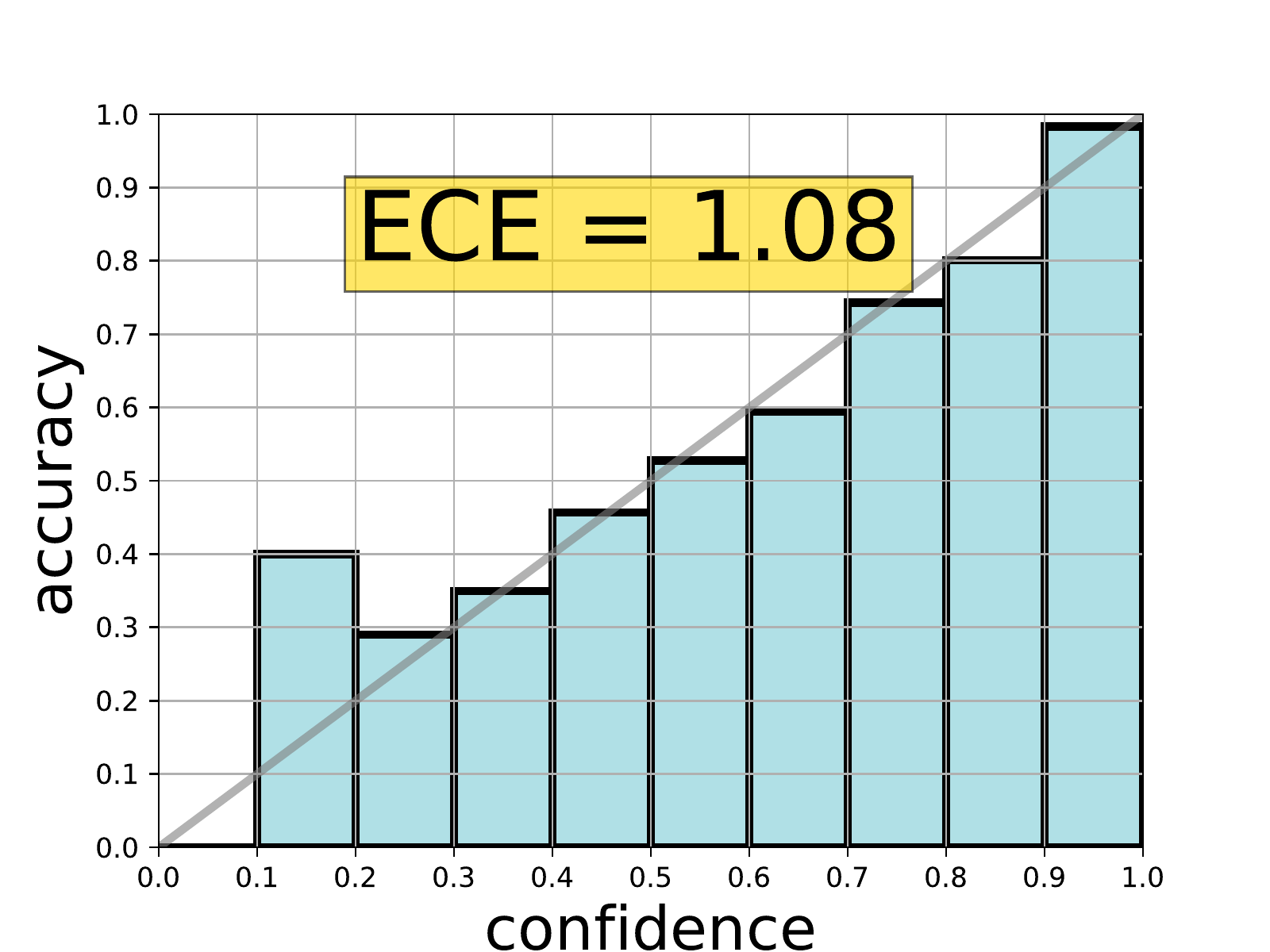}
		\caption{CARING-I3D, \\ \null \quad common classes}
		\label{fig:i3d_caring_common}
	\end{subfigure}%
	\begin{subfigure}{.16\textwidth}
		\centering
		\includegraphics[trim={0.1cm 0 1.4cm 0},clip,width=\linewidth]{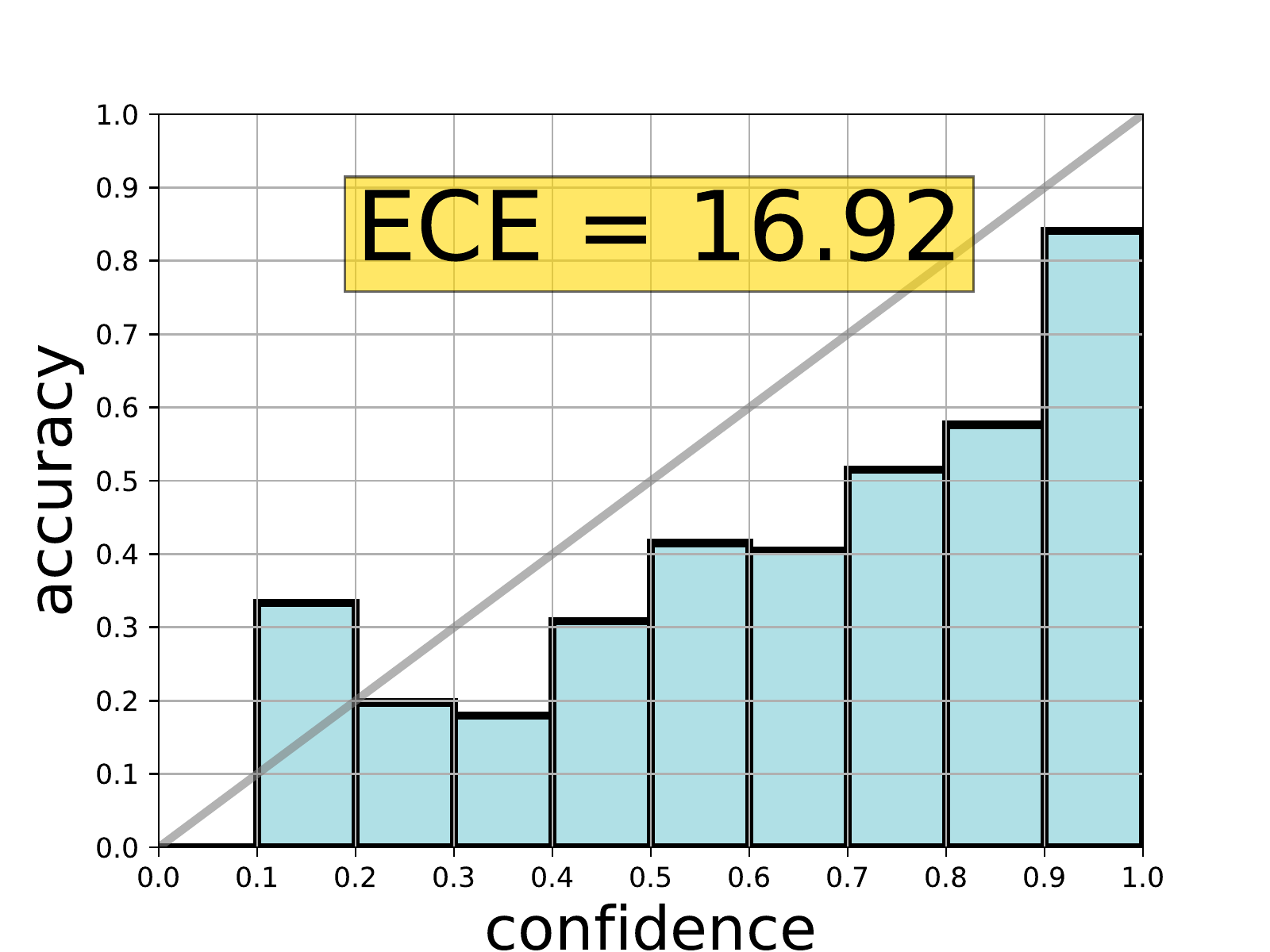}
		\caption{P3D (original), \\ \null \quad common classes}
		\label{fig:p3d_common}
	\end{subfigure}%
	\begin{subfigure}{.16\textwidth}
		\centering
		\includegraphics[trim={0.1cm 0 1.4cm 0},clip,width=\linewidth]{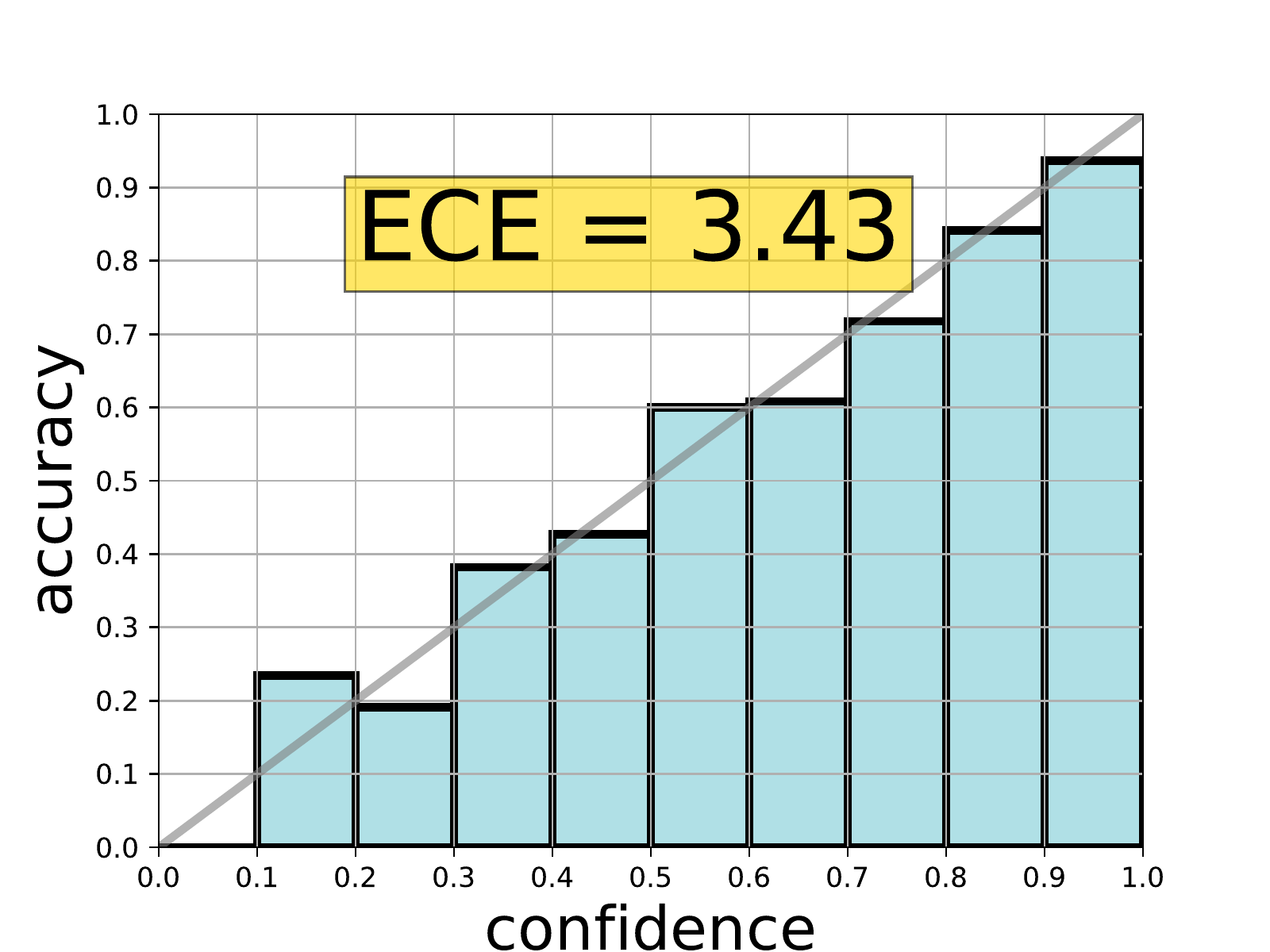}
		\caption{P3D + temp. scaling, common classes}
		\label{fig:p3d_temp_common}
	\end{subfigure}%
	\begin{subfigure}{.16\textwidth}
		\centering
		\includegraphics[trim={0.1cm 0 1.4cm 0},clip,width=\linewidth]{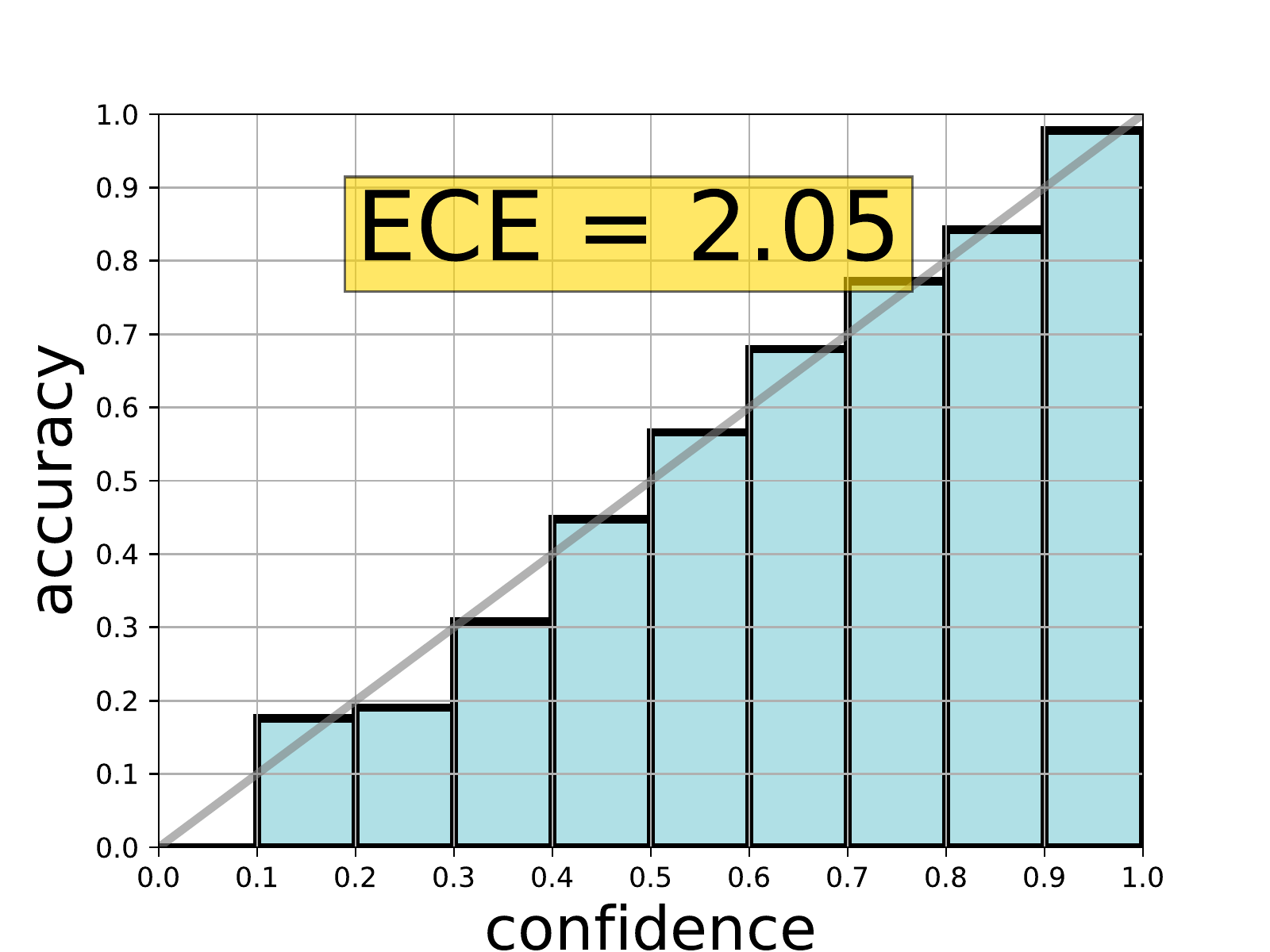}
		\caption{CARING-P3D, \\ \null \quad common classes}
		\label{fig:p3d_caring_common}
	\end{subfigure}%
	
	\begin{subfigure}{.16\textwidth}
		\centering
		\includegraphics[trim={0.1cm 0 1.4cm 0},clip,width=\linewidth]{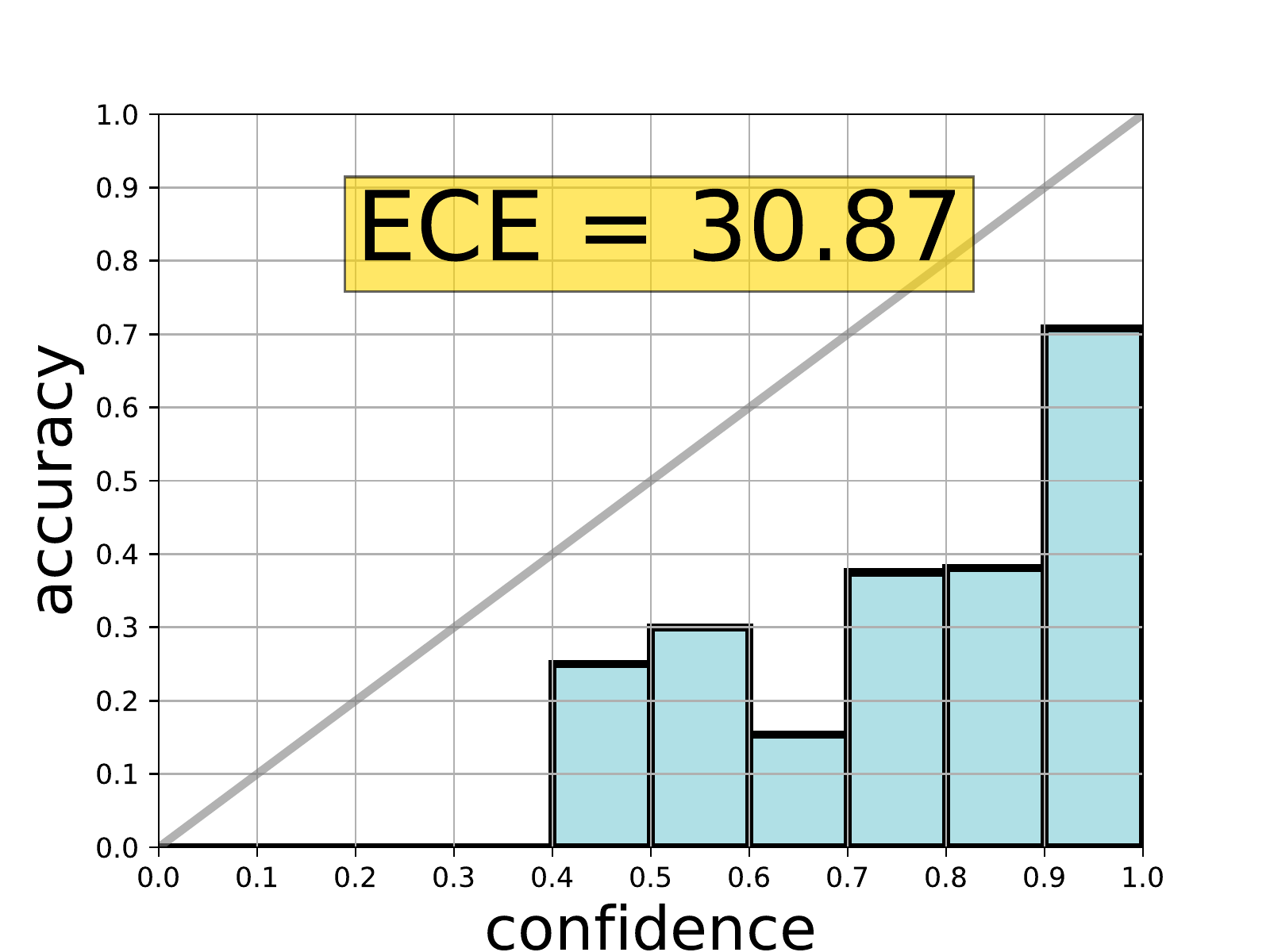}
		\caption{I3D (original), \\ \null \quad rare   classes}
		\label{fig:i3d_rare}
	\end{subfigure}%
	\begin{subfigure}{.16\textwidth}
		\centering
		\includegraphics[trim={0.1cm 0 1.4cm 0},clip,width=\linewidth]{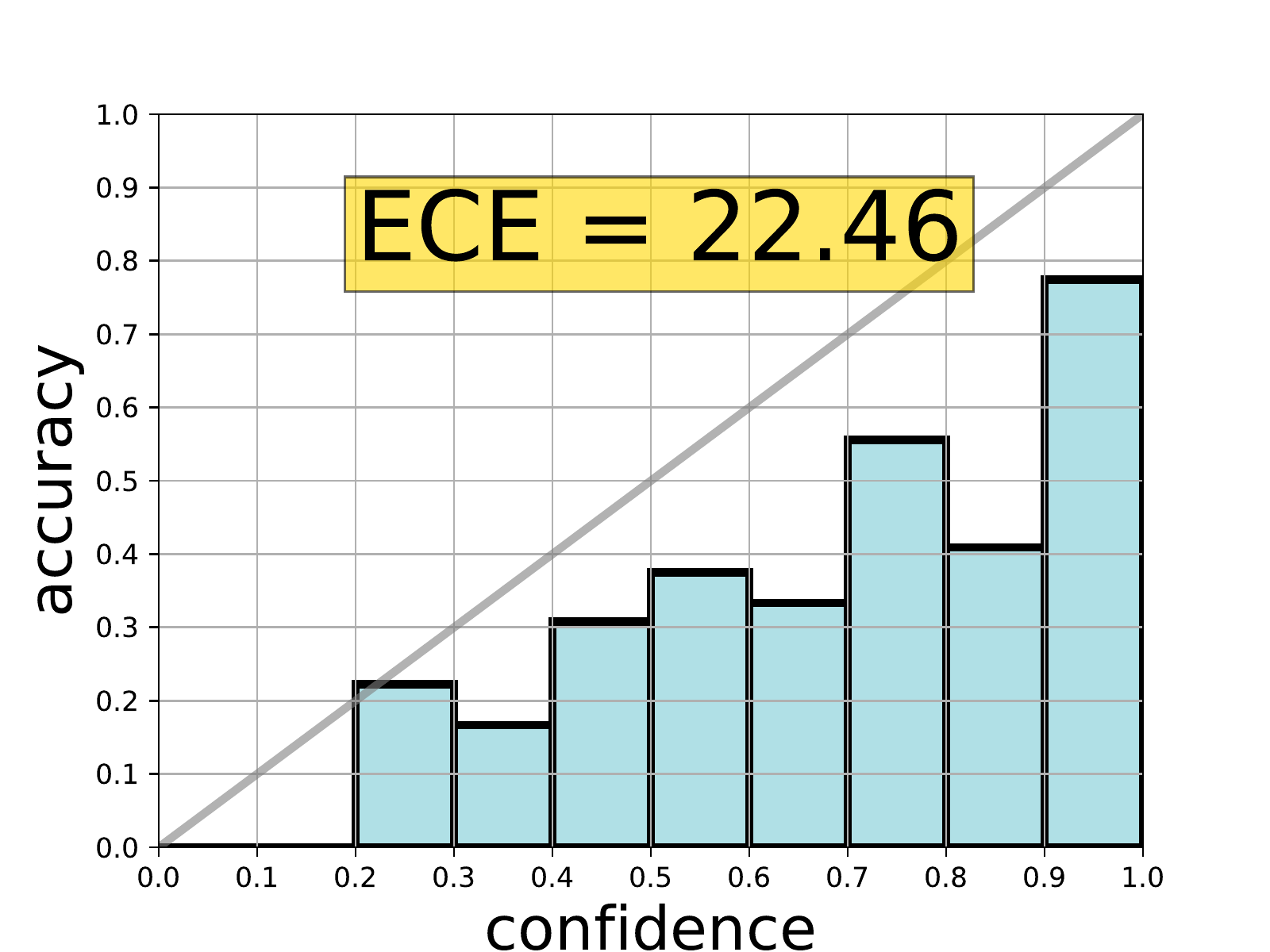}
		\caption{I3D + temp. scaling, rare  classes}
		\label{fig:i3d_temp_rare}
	\end{subfigure}%
	\begin{subfigure}{.16\textwidth}
		\centering
		\includegraphics[trim={0.1cm 0 1.4cm 0},clip,width=\linewidth]{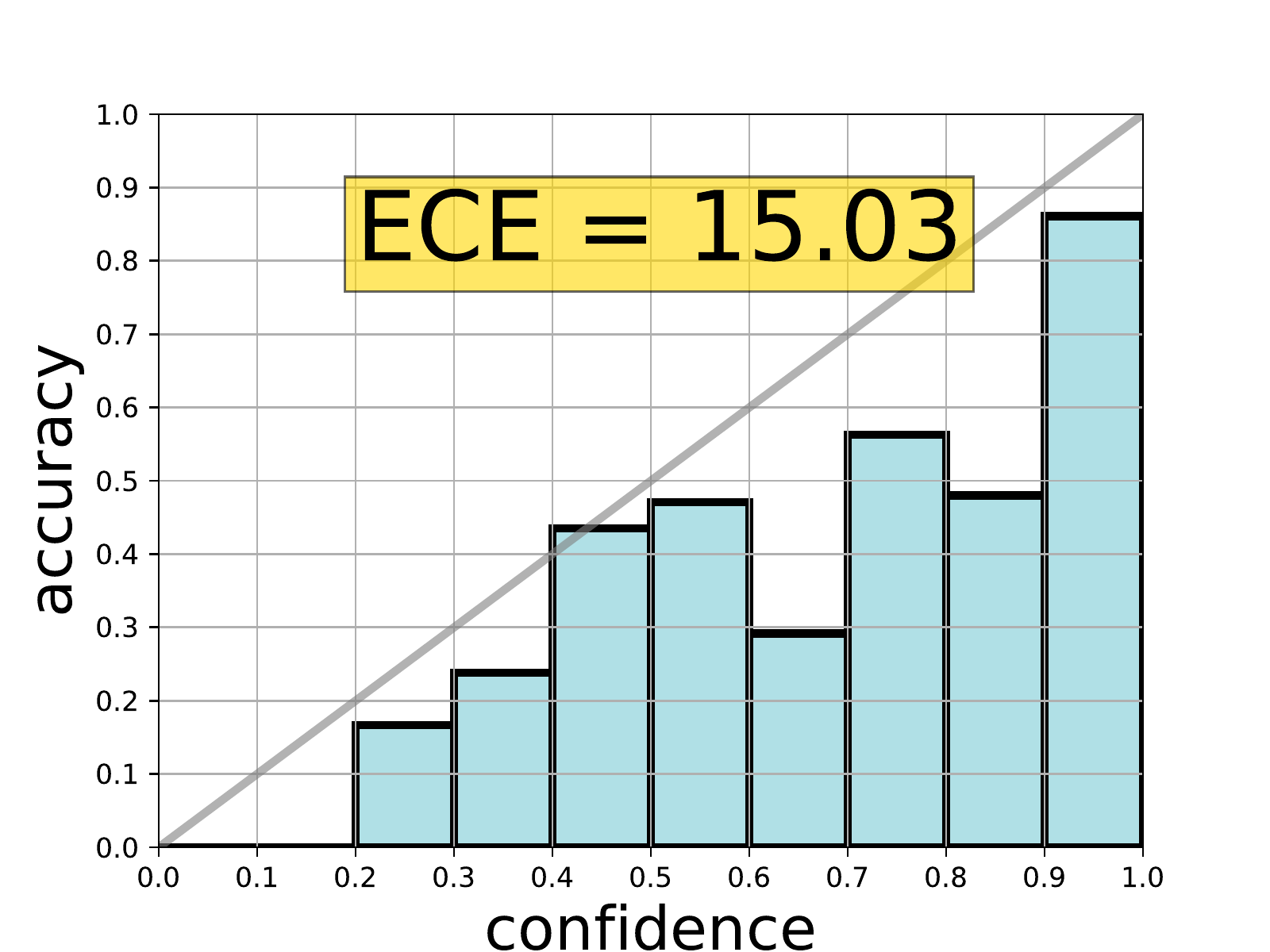}
		\caption{CARING-I3D, \\ \null \quad rare  classes}
		\label{fig:i3d_caring_rare}
	\end{subfigure}%
	\begin{subfigure}{.16\textwidth}
		\centering
		\includegraphics[trim={0.1cm 0 1.4cm 0},clip,width=\linewidth]{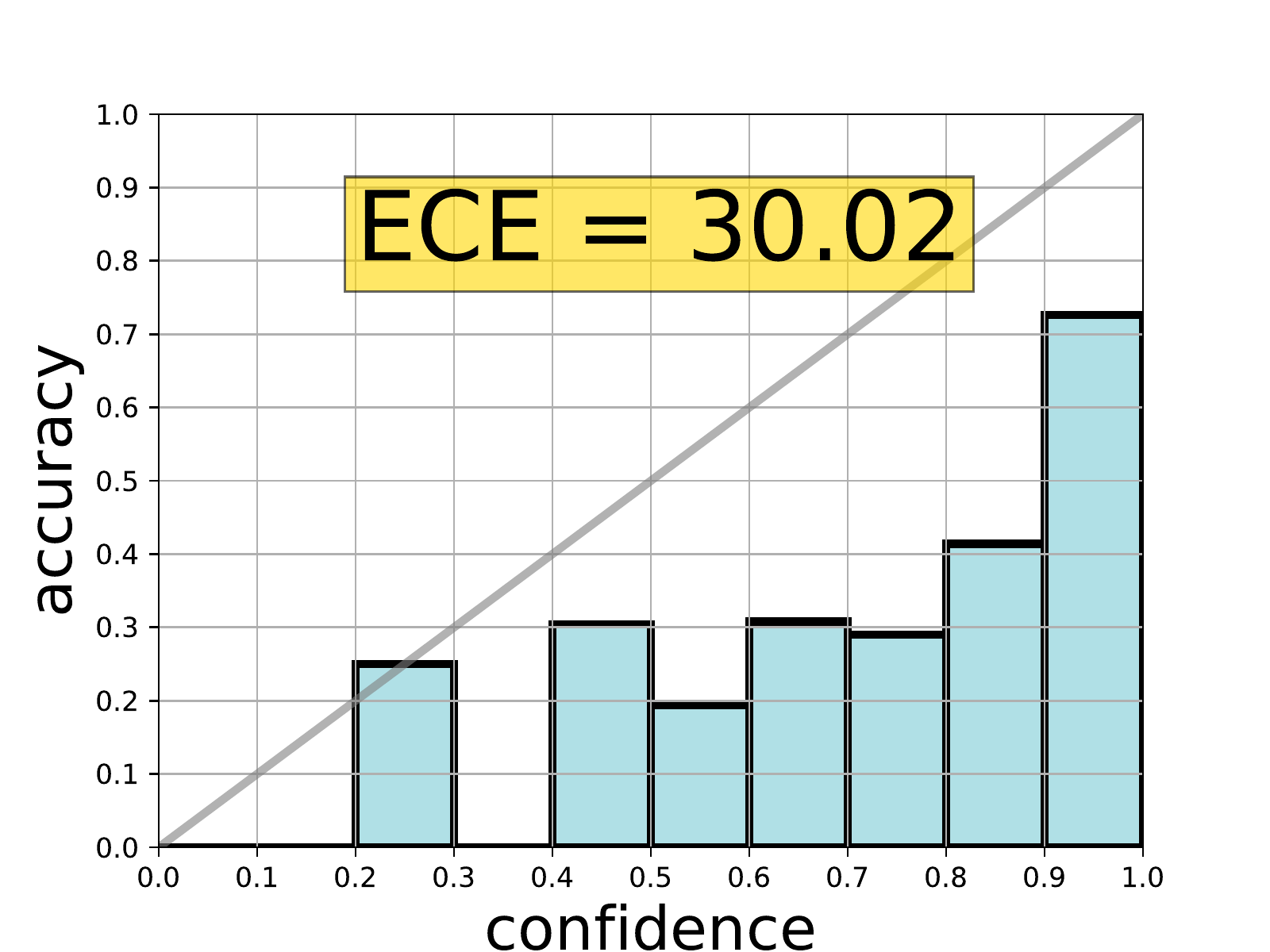}
		\caption{P3D (original), \\ \null \quad rare  classes}
		\label{fig:p3d_rare}
	\end{subfigure}%
	\begin{subfigure}{.16\textwidth}
		\centering
		\includegraphics[trim={0.1cm 0 1.4cm 0},clip,width=\linewidth]{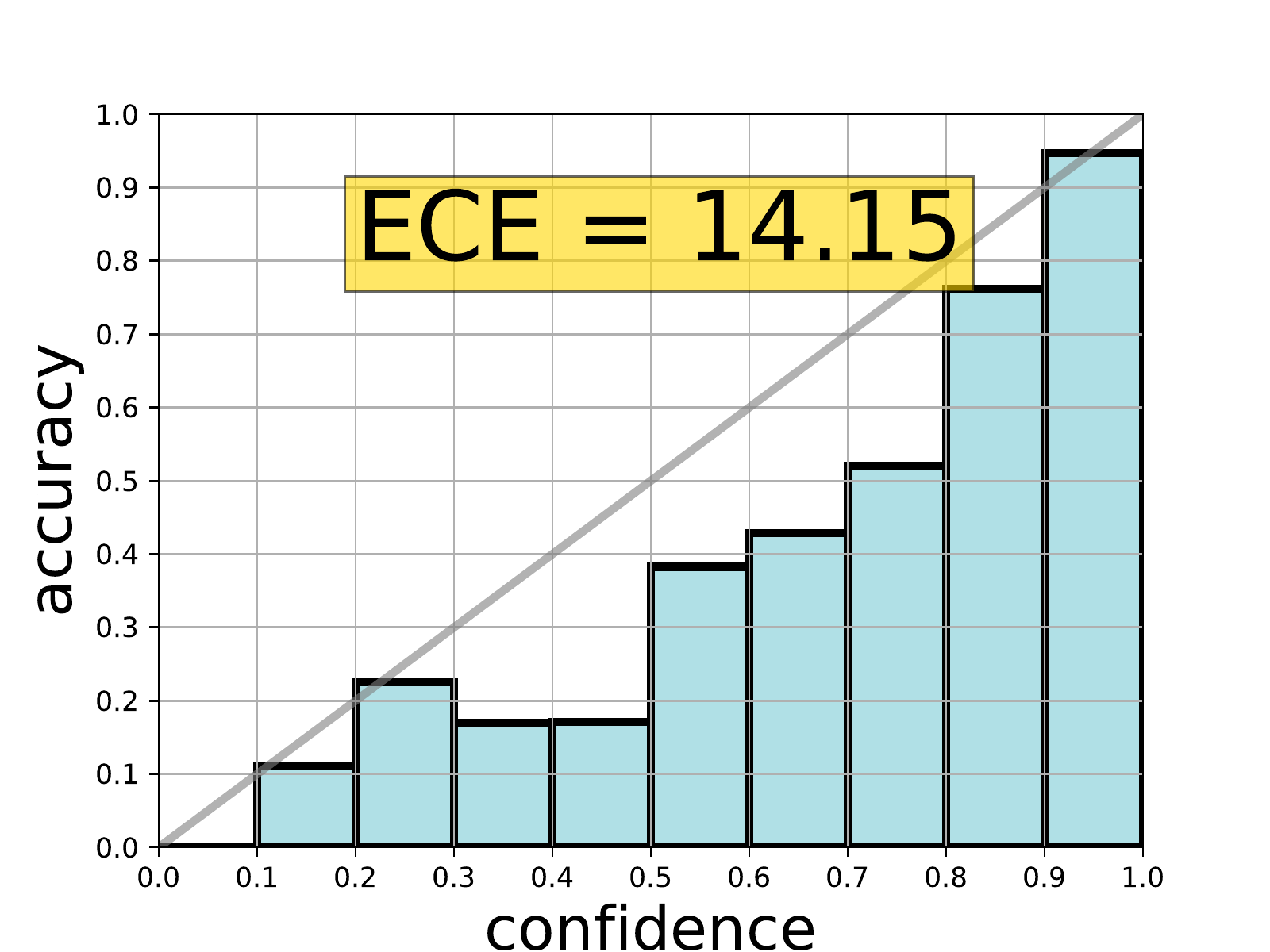}
		\caption{P3D + temp. scaling, rare  classes}
		\label{fig:p3d_temp_rare}
	\end{subfigure}%
	\begin{subfigure}{.16\textwidth}
		\centering
		\includegraphics[trim={0.1cm 0 1.4cm 0},clip,width=\linewidth]{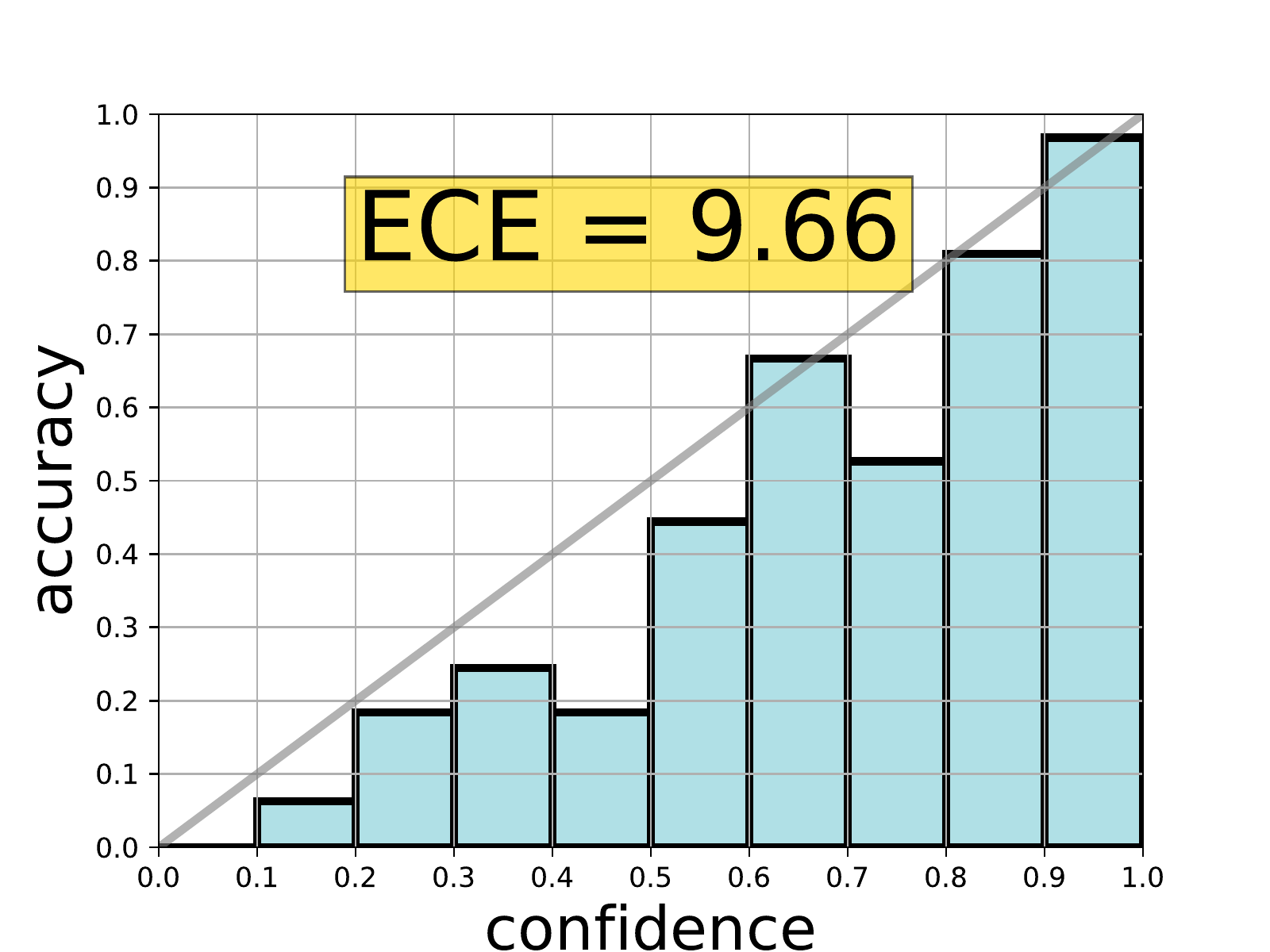}
		\caption{CARING-P3D, \\ \null \quad rare  classes}
		\label{fig:p3d_caring_rare}
	\end{subfigure}%
	\caption{Reliability diagrams of different models reflect the agreement between the confidence values and the empirically measured probability of  correct prediction (results of one Drive\&Act validation split). A model with \emph{perfectly calibrated uncertainty scores would match the diagonal} (a detailed explanation in Section \ref{sec:calib_diag} ). Note, that the ECE values deviate from Table \ref{tbl:reliability_results}, as they visualize a single split, while the final reported results are averaged over all splits.
		While the temperature scaling consistently improves the confidence estimates, our CARING model leads to the lowest calibration error in all settings. } 
	\label{fig:reliability_diag}
	\vspace{-0.1cm}
\end{figure*}

We now take a closer look at how the obtained probability estimates are distributed.
Figure \ref{fig:density_confidence} plots these distributions, where the color highlights the frequency of \textit{Drive\&Act} validation set samples obtaining a certain confidence score for correct and incorrect predictions. 
The X axis denotes the improved probability estimates after the CARING transformation, while the Y  axis corresponds the original \textit{Softmax} confidences.
For correct predictions (Figure \ref{fig:density_confidence} on the left), model confidences tend to be high both with and without CARING, but the CARING confidences have much higher variance, which is visible through the horizontal spread of the density plot.
This effect is even stronger in case the model made a mistake (Figure \ref{fig:density_confidence} on the right): the red high-frequency region is much wider  on the X-axis than on the Y-axis, highlighting more reserved confidence values for CARING.

Figure~\ref{fig:reliability_diag} visualizes the agreement between the estimated model confidence and the empirically measured probability of the correct outcome by using \textit{reliability diagrams} (explained in Section~\ref{sec:definition}).
In case of good estimates, the result will be closely approaching the diagonal line.
Values above the diagonal are correlated to models being over-confident in their prediction, while values below imply that the model harbours suspicions about the outcome and the accurate prediction probability is higher than assumed.

We first analyze the reliability diagrams of the original action recognition networks.
Both P3D and I3D confidence values largely deviate from the target, showing clear biases towards excessively optimistic scores (\ie, values are oftentimes below the diagonal in Figures~\ref{fig:i3d_all}, \ref{fig:p3d_all}, \ref{fig:i3d_common}, \ref{fig:p3d_common}, \ref{fig:i3d_rare}, \ref{fig:p3d_rare}).
One exception worth discussion is an above-diagonal peak in the low probability segment for \emph{all} and \emph{common} classes, hinting that in ``easier'' settings, low confidence samples often turn out to be correct (\ref{fig:i3d_all}, \ref{fig:p3d_all}, \ref{fig:i3d_common}, \ref{fig:p3d_common}).
Still, in the ``harder'' setting of \emph{rare} activities (Figure \ref{fig:i3d_rare}, \ref{fig:p3d_rare}), the biases towards too high probabilities are noticeable for all values.

We have observed a clear positive effect of temperature scaling (Figures \ref{fig:i3d_temp_all}, \ref{fig:p3d_temp_all}, \ref{fig:i3d_temp_common}, \ref{fig:p3d_temp_common},  \ref{fig:i3d_temp_rare}, \ref{fig:p3d_temp_rare})
and our CARING model (Figures \ref{fig:i3d_caring_all}, \ref{fig:p3d_caring_all}, 
\ref{fig:i3d_caring_common},  \ref{fig:p3d_caring_common},  \ref{fig:i3d_caring_rare}, \ref{fig:p3d_caring_rare}).
Evidently, CARING models outperform other approaches in all settings and lead to almost perfect reliability diagrams for \emph{all} and \emph{common} classes.
Yet, both temperature scaling and CARING strategies still have issues in dealing with \emph{rare} classes, with confidence being too high, indicating a vital direction for future explorations.

\begin{figure*}[!t]
	\centering
	\includegraphics[width=0.95\linewidth ]{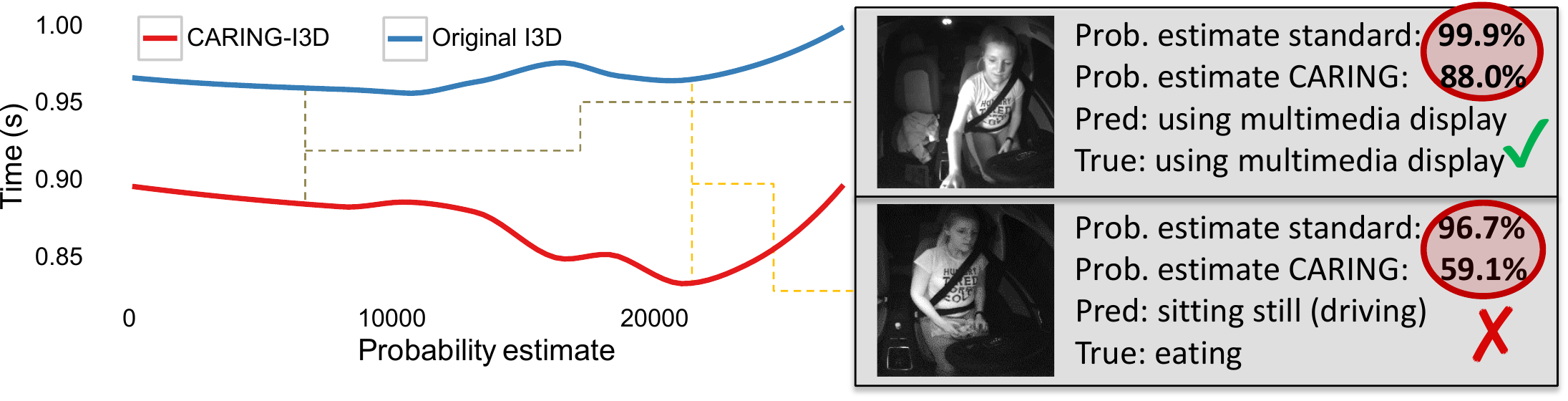}
	\caption{\revised{Example of average confidence estimates (median value with a sliding window range of one minute) during a single driving session  for the original (blue line) and CARING (red line) models and qualitative examples of a correct and an incorrect outcome (on the right). }}
	\label{fig:single_session}
		\vspace{-0.3cm}
\end{figure*}

It is worth noting that ECE might be in a slight disarray with the visual reliability diagram representation, as the metric weighs the misalignment in each bin by the amount of data points in it, whereas the reliability diagrams do not manifest such frequency distribution.
For instance, while the CARING-I3D model in Figure~\ref{fig:i3d_caring_common} marginally exceeds the target diagonal, it has lower expected calibration error than CARING-P3D which seems to yield nearly perfect results in Figure~\ref{fig:p3d_caring_common}.
As there are only very few examples in the low-confidence bin, they are overshadowed by smaller differences in the high-confidence bins, which therefore contribute more to the final.

\section{Discussion: Towards Reliable and\\ Secure Driver Observation}

\revised{\subsection{Significance for Driver Observation} }

\revised{
Although CARING consistently improves the realism of confidence values, the results vary greatly depending on the behaviour type (Figure \ref{fig:results_all_classes}).
As it comes to real-life applications inside the vehicle, the maturity level of the proposed model depends on the concrete use-case. 
Drive\&Act is a  fine-grained dataset with both,  manual  and fully automated driving, and therefore covers a high number of secondary activities. 
Some of these behaviours were almost unthinkable behind the steering wheel until now, but will become more common in the future (such as \textit{working on laptop}). 
For the use-case of identifying driver distraction, the most significant Drive\&Act  category is \textit{sitting still}, which comprises both, (1) driving  and (2) sitting with the hands not on the steering wheel, while the car is in the self-driving mode (without any distractive activities).
This category is inherently linked to lower distraction  and distinguishing it from other driver states is crucial for identifying distraction at SAE levels 0 to 3.
Recognition of this behaviour, as well as most of other behaviours prevalent in manual driving, (\textit{e.g.}, \textit{using multimedia display}, \textit{fastening seat belt}) consistently yields high-quality confidence estimates.
The rather long-term use-case of increased comfort through automatic driver-centered adaptation, (\textit{e.g.}, softer driving if the person is drinking tea or sleeping at SAE levels 4 and 5), on the other hand, might require  finer granularity of driver behaviours.
In this case, the readiness level of our system strongly depends on the concrete application and significant improvement would still be required if very concise recognition  is desired (for example, \textit{opening} vs. \textit{closing bottle}). 
Nevertheless, the most fundamental categories, such as \textit{eating},  \textit{reading} or \textit{interacting with phone} achieve ECE $<10\%$, which might be sufficient for most of the applications.
}

\revised{
Our experiments indicate that  previously used driver observation models do not provide proper uncertainty estimates, therefore ignoring their own limits.
Figure \ref{fig:single_session} illustrates an example of average confidence estimates during a single driving session of the Drive\&Act dataset for the original (blue line) and CARING (red line) models. 
The original model consistently stays in the area $>95\%$, indicating strong overconfidence, since the average Drive\&Act accuracy is clearly below this value.
Two qualitative examples on the right showcase very high confidence values (99.9\% and 96.7\%) of the original model in cases of both, correct predictions and misclassifications.
Such overly confident models provide a false sense of security and are dangerous if affecting the overall driving dynamics.
Furthermore, such false-positives predictions are very damaging for the driving experience and human-vehicle interaction. 
In the vast majority of cases, the uncertainty error was towards overconfidence, which is clearly improved though our CARING model. 
In the example of \ref{fig:single_session}, CARING  clearly lowers the average confidence estimate (red line).
In the qualiative examples on the right, CARING estimate is much lower in case of misclassification ($59.1\%$ vs. $88.0\%$). 
Still, one might argue that a value of $59.1\%$ is relatively high and finding a good threshold for acceptance of the recognition result remains an open question.
}

\revised{
Classification uncertainty plays a significant role in driver observation, as the the models are deployed in a dynamic world where domain shifts, novel behaviours and other unforeseen situations may occur at any time. 
Besides being potentially dangerous, false-positives caused by overly confident models are often highly disturbing for the user. For example, in the use-case of  driver-centered adaptation, repeated false recognition that the human is, \textit{\eg}, reading triggering a reaction like turning on the light would strongly bother the user.
Apart from the direct benefits of proper confidence values for decision-making systems, good assessment of uncertainty enhances model interpretability. 
For example, in the realistic scenario of open-world recognition, low-confidence input might be passed to human experts, which would provide the correct annotations (\ie~active learning) and therefore improve the decision boundary.
In summary, obtaining realistic uncertainty estimates is important in real-life driving systems due to (1) safety, (2) better human-machine interaction (through avoidance of false positives triggering an unwanted response), and (3) better interpretability. 
}

\subsection{Discussion of Limitations}

While our method leads to much more reliable confidence estimates in driver behaviour analysis, it is not without limitations. 
First, we observe, that the accuracy-confidence mismatch is still relatively large in case of activities with little training examples. 
Secondly, while we achieve the best results with the proposed CARING approach, we acknowledge, that this comes with increased computational cost due to an additional external calibration network.
This overhead of course strongly depends on the architecture choice and is comparably small in our case ($66.7 K$ Parameters and $0.13$ MFLOPS for our two-layer calibration network using $1024$ dimensional input representation vector holding the intermediate I3D representations).
Furthermore,  \emph{identifying misclassifications}  among the training classes is not the same as \emph{identifying novelty}~\cite{ovadia2019can}.
The essence of  calibration-based methods, such as temperature scaling or our \textit{CARING} model, is learning  realistic confidence values on a held-out validation set. 
As shown in a recent study from the area of image classification~\cite{ovadia2019can}, the reliance on this held-out validation set also becomes the greatest weakness of such models when facing domain shifts.
In other words, confidence calibration is effective, as long as the test data roughly reflects the distribution of the validation set. 
Producing probability estimates of the observed behavior class that was previously unseen during training is therefore a different challenge that we aim to address in the future.
Nevertheless, our work  takes a step towards driver observation models which are not only  accurate, but  also to determine, how likely they are to be right in their prediction, which is vital for reliable and transparent  decision systems within the vehicle.

\subsection{Outlook: Reliable and Secure Driver Observation}

The goal of driver observation is to delegate more actions and decisions to the vehicle itself and reacting to the captured driver states accordingly.  
As in any decision process, the reliability and security of the incoming information become important. 
Our current work provides an accurate method for measuring the uncertainty of this information. 
Taking this as a  starting point, one could follow different research directions which are currently rather overlooked.

\textit{Adversarial attacks} are one of such  examples. 
Though thoroughly explored for signals coming from the exterior of the vehicle~\cite{adversarial_attacks_and_defenses,Li2021AdaptiveSA}, it is still not known how extensive the risk is of an attack against the cameras inside the vehicle cabin. 
Unfortunately, deep learning research suggests that raw Softmax confidences are highly unreliable in case of adversarial attacks~\cite{nguyen2015deep}.
Analyzing the uncertainty of the networks can reduce the effect of such attacks and provide a way to detect them. 
Still, how much additional effort is indeed needed to penetrate this second layer of protection for action recognition models remains an open question.

A related direction is that of \textit{interpretability}~\cite{zablocki2021explainability}. 
As we move towards end-to-end approaches, it is easier for dataset biases to influence predictions. 
In the case of driver observation, a movement of the hand towards the mouth could be for example misinterpreted as eating~\cite{Roitberg2020_InterpretableCNN}. 
Visualizing the features which determine the predictions can help build trust and ensure that situations as the aforementioned will not occur. 
The CARING model allows for a second interpretation, namely that of the uncertainty estimates which could lead to additional insights into model reasoning.

Last but not least comes \textit{security and privacy}~\cite{privacy_liability,security_of_autonomous_driving}. Monitoring the inside of a vehicle exposes a lot of private sensitive information. Aside form the video feed itself, the processed information is also of high concern, especially when used to evaluate professional drivers, resolve accident disputes, or used for marketing purposes. In all those cases, knowing the reliability of predictions can help to better quantify the benefits and dangers of using the driver observation information.

\section{Conclusion}

While conventional driver observation research targets high-accuracy frameworks in the first place, our goal is to develop driver activity recognition models which are not only accurate, but also produce  uncertainty estimates that can safely be interpreted as probabilities of a correct outcome.
First, we study the reliability of model confidences of existing driver activity recognition models and come to worrying conclusions: there is a significant discrepancy between the original confidence estimates of the networks and the empirically measured probabilities of a correct prediction.
To remedy this, we implement two methods for calibrating the network estimates:  leveraging the temperature scaling approach often used in image recognition for such purposes and our newly proposed model, which we refer to as CARING.
Our CARING model learns to dynamically generate values used for scaling the model confidences based on input representation with an external calibration network.
Our extensive experiments indicate strong improvement through temperature scaling alone, while our CARING model consistently \revised{yields the best results}.
\revised{Our study provides} encouraging evidence, that the proposed uncertainty-aware models can be efficiently used to develop reliable driver observation systems.

\bibliography{bib}{}
\bibliographystyle{IEEEtran}


\end{document}